\documentclass[11pt]{article}
\usepackage[margin=1in]{geometry}
\usepackage{amsmath,amssymb}
\usepackage[utf8]{inputenc}
\usepackage[T1]{fontenc}
\usepackage{lmodern}
\usepackage{textcomp} 
\usepackage{graphicx}
\usepackage[square,numbers]{natbib}
\usepackage{hyperref}
\usepackage{url}
\usepackage{listings}
\usepackage{calc} 
\usepackage{subcaption} 
\usepackage{longtable}
\usepackage{booktabs}
\usepackage{array}
\usepackage{float}
\usepackage{xcolor}

\providecommand{\tightlist}{%
  \setlength{\itemsep}{0pt}\setlength{\parskip}{0pt}}

\DeclareUnicodeCharacter{00B0}{\textdegree} 
\DeclareUnicodeCharacter{2011}{-} 
\DeclareUnicodeCharacter{2013}{--} 
\DeclareUnicodeCharacter{2014}{---} 
\DeclareUnicodeCharacter{2212}{\ensuremath{-}} 
\DeclareUnicodeCharacter{2248}{\ensuremath{\approx}} 
\DeclareUnicodeCharacter{2260}{\ensuremath{\neq}} 
\DeclareUnicodeCharacter{2264}{\ensuremath{\le}} 
\DeclareUnicodeCharacter{2265}{\ensuremath{\ge}} 
\DeclareUnicodeCharacter{0394}{\ensuremath{\Delta}} 
\DeclareUnicodeCharacter{03B5}{\ensuremath{\varepsilon}} 
\DeclareUnicodeCharacter{27F7}{\ensuremath{\longleftrightarrow}} 

\providecommand{\real}[1]{#1}

\definecolor{codebackground}{rgb}{0.95,0.95,0.95}
\definecolor{codeborder}{rgb}{0.8,0.8,0.8}
\title{Benchmarking Large Language Models for Geolocating Colonial Virginia Land Grants*}
\author{Ryan Mioduski, Independent Researcher}
\date{2025-07-23}

\begin{document}
\maketitle

\renewcommand{\thefootnote}{\fnsymbol{footnote}}
\footnotetext[1]{Code and data: \url{https://github.com/ryanmio/colonial-virginia-llm-geolocation} • Zenodo DOI: \url{https://doi.org/10.5281/zenodo.16269949}}
\renewcommand{\thefootnote}{\arabic{footnote}}

\section{Introduction}\label{introduction}

\subsection{Historical Context \&
Motivation}\label{historical-context-motivation}

Virginia's colonial land patents are a cornerstone resource for scholars
studying settlement patterns, the political economy of plantation
agriculture, and Indigenous dispossession in the seventeenth and
eighteenth centuries. Yet the spatial dimension of these sources remains
under-exploited: most patents survive only as narrative metes-and-bounds
descriptions in printed abstract volumes such as \emph{Cavaliers and
Pioneers} (C\&P) \citep{Nugent1979_cavaliers3}. Without geographic
coordinates, historians and archaeologists cannot readily visualize how
land ownership evolved or test hypotheses with modern Geographic
Information System (GIS) tools. Creating a machine-readable,
georeferenced version of C\&P would unlock new quantitative approaches
to long-standing questions about colonial Virginia's social and
environmental history.

Digitizing and geo-locating the abstracts, however, is notoriously
labor-intensive. Even professional GIS analysts can spend several hours
per grant reconciling archaic place-names, inconsistent spellings, and
low-resolution boundary calls. Recent breakthroughs in large language
models (LLMs) suggest a new pathway: language-driven spatial reasoning
where a model reads the patent text and predicts latitude/longitude
directly or with minimal tool assistance. This study explores whether
current-generation LLMs can shoulder that burden accurately and cheaply
enough to matter for digital history.

\begin{quote}
Example grant abstract (C\&P Vol. 3, pp.~78--79; abstracting Patent Book
9, p.~571; 23 Oct.~1703):

``JOHN POYTHRESS, 609 A., 2 R., \& 9 P., Chas. City Co; on S. side of
James River; Beg. on S. side the Black Water; to the Nottoway Path; to
the Black Water Spring; along the Sw.; near Capt.~Robert Lewcy; by
Townes' Quarter; to Hercules Flood; taking in a point containing 50 acs;
\ldots{}''
\end{quote}

This typical abstract uses abbreviated surveyor jargon (e.g., Sw., Riv.,
br.), chained landmarks, and adjacency clauses to neighboring tracts to
narratively describe the parcel.

\subsection{Problem Statement}\label{problem-statement}

Despite the promise of LLMs, their ability to extract usable coordinates
from early-modern archival prose had not been systematically evaluated
prior to this work. Key uncertainties addressed in this study included:

\begin{itemize}
\tightlist
\item
  Could large language models trained mostly on contemporary text
  reliably parse seventeenth-century toponyms and bearing conventions
  and resolve them to coordinates?
\item
  Would providing API-based tools (e.g., Google Places search)
  materially improve accuracy relative to a pure text approach?
\item
  How did model predictions compare to single-analyst GIS workflow
  \citep{Bashorun2025_gis}, deterministic pipelines such as the GeoTxt
  Stanford NER geoparser \citep{karimzadeh2019geotxt}, neural geoparsers
  like Mordecai-3 \citep{halterman2023mordecai}, and other heuristic
  benchmarks in both error and cost?
\end{itemize}

Addressing these questions required a rigorously annotated test bench
that blended historical sources, modern GIS ground truth, and controlled
prompt engineering. The methodological design seeks to embody principles
of rigorous and responsible GeoAI research, as outlined by Li et
al.~\citep{Li2024_geoai} and detailed further in Section 2.4,
and---relative to classic geoparsers such as Stanford NER/GeoTxt
\citep{karimzadeh2019geotxt}, Mordecai‑3 \citep{halterman2023mordecai},
and the map‑aware CamCoder \citep{Gritta2018_camcoder}---centers on
long, archaic abstracts with nested landmarks and surveyor bearings
rather than the short, contemporary texts common in prior benchmarks.
Accordingly, we compare a pure text‑only pipeline to an LLM‑plus‑tools
variant that can call geocoders mid‑reasoning, and we report accuracy,
monetary cost, and latency as first‑class outcomes to situate our study
within modern LLM geolocation while targeting a distinct colonial corpus
and evaluation protocol.

\subsection{Contributions}\label{contributions}

This study makes four principal contributions:

\begin{enumerate}
\def\labelenumi{\arabic{enumi}.}
\tightlist
\item
  Releases the first copyright-compliant, machine-readable dataset of
  \emph{Cavaliers and Pioneers}, Vol. 3 \citep{Nugent1979_cavaliers3},
  including (i) row-level metadata---row identifier, word count, and
  SHA-256 hash---for all 5,471 abstracts, and (ii) limited,
  non-substitutable excerpts of up to 200 words for the 43 evaluation
  abstracts. The complete OCR text (5\,471 abstracts) has been archived
  on Zenodo (DOI: 10.5281/zenodo.16269949); qualified researchers can
  download it for non‑commercial research.
\item
  Provides authoritative latitude/longitude pairs for 43 randomly
  sampled patents, derived from GIS polygons created by the nonprofit
  project One Shared Story (OSS) from public-domain archival sources and
  cross-validated by scholars, yielding a high-fidelity evaluation
  target for this benchmark.
\item
  Presents the first systematic benchmarking of large language models on
  historical land-grant geolocation, evaluating two prompting
  paradigms---direct-to-coordinate inference and tool-augmented
  chain-of-thought---across six OpenAI models spanning the o-series,
  GPT-4-class, and GPT-3.5 architectures, including detailed ranking of
  model accuracy, inference costs, and latency.
\item
  Quantifies trade-offs among spatial error, monetary expense,
  processing time, cost, and latency, demonstrating that a pure LLM
  pipeline can match or surpass a single-analyst GIS workflow, Stanford
  NER geoparser, Mordecai-3 neural geoparser, and county-centroid
  heuristic, while operating substantially faster and more
  cost-effectively in this 43-grant pilot evaluation.
\end{enumerate}

All data, code, and results are available in the supplemental
repository:
\url{https://github.com/ryanmio/colonial-virginia-llm-geolocation}; the
raw corpus and evaluation data are also archived on Zenodo (DOI:
10.5281/zenodo.16269949).

\section{Background \& Related Work}\label{background-related-work}

\subsection{Historical GIS and Land-Grant
Mapping}\label{historical-gis-and-land-grant-mapping}

Digitizing colonial-era land grants has long promised new insights into
European settlement patterns, Indigenous land displacement, and the
development of local economies. However, this potential has been
constrained by the extensive manual labor required to convert
metes-and-bounds descriptions into spatial data. Traditional approaches
to georeferencing these historical records have proven prohibitively
time-consuming - a genealogical case study by Julian and Abbitt
\citep{Julian2014_tennessee} required nearly ten years of archival
sleuthing and three university-semester GIS projects to pinpoint a
single family's land claims across three Tennessee counties.

Several institutional efforts have attempted to address these
challenges, though coverage remains incomplete. The Library of Virginia
maintains a statewide \emph{Land Patents and Grants} online database
hosting scanned images and searchable indices for every recorded patent
(1623--1774) and subsequent grant (1779--2000), including Northern Neck
surveys, but provides no ready-made GIS polygons, limiting its direct
utility for spatial analysis \citep{lva_patents_db}. Similarly, Loudoun
County GIS staff have successfully reconstructed all original grants
within their jurisdiction \citep{loudoun_grants_dataset}. These
initiatives demonstrate the feasibility of digitizing historical land
records but also highlight significant gaps in existing datasets - many
seventeenth- and eighteenth-century patents still lack spatial
coordinates.

Among the most thorough academic efforts for Virginia's Northern Neck
proprietary are Mitchell's \citep{mitchell1977whiteoak} maps and
companion text documenting the ``Beginning at a White Oak'' patents of
Fairfax County. This work reconstructed hundreds of early land grants
with polygonal boundaries, establishing both the feasibility and
research value of transforming metes-and-bounds descriptions into
spatial data. Building on such foundations, scholars have leveraged
available georeferenced grants for substantive historical analysis. In
Virginia, seminal studies like Fausz \citep{Fausz1971_settlement}
utilized narrative patent abstracts to trace settlement patterns along
the James River basin, while noting the persistent challenges of
transforming textual descriptions into precise spatial coordinates for
quantitative analysis.

This analytical potential extends beyond Virginia. Dobbs
\citep{Dobbs2009_backcountry} used georeferenced North Carolina grants
to demonstrate that eighteenth-century town sites often followed
pre-existing Indigenous trails, while Coughlan and Nelson
\citep{Coughlan2018_settlement} leveraged a dataset of 1,160 South
Carolina grants to model settlement patterns based on river access and
soil analysis. In each case, spatial enablement of historical records
revealed patterns difficult to discern through textual sources alone.

In genealogical and historical research communities, semi-automated
solutions have emerged to assist with this labor-intensive process.
DeedMapper software \citep{DeedMapper_software} helps researchers
convert metes-and-bounds descriptions into visual plots, though it still
requires manual entry of deed text and expert positioning of parcels on
reference maps. Professional development courses from the Salt Lake
Institute of Genealogy (SLIG) continue to teach these specialized
mapping techniques, reflecting the still-developing state of automation
in this field.

The literature establishes three critical facts. First, historians value
land-grant GIS layers because they unlock settlement and landscape
questions that text alone cannot answer. Second, traditional platting
methods are too slow and too localized to deliver colony-scale coverage.
Third, the piecemeal datasets that do exist furnish both ground truth
and a methodological benchmark for any attempt at automation. This study
addresses this bottleneck by testing whether large language models can
shoulder the coordinate-extraction burden---potentially transforming
Virginia's colonial patents from archival prose to research-ready GIS at
scale.

\subsection{Large Language Models for
Geolocation}\label{large-language-models-for-geolocation}

Building on the manual coordinate-extraction bottleneck outlined in
Section 2.1, recent advances in large language models (LLMs) suggest
that much of the geoparsing pipeline can now be automated. Coordinate
extraction---sometimes called \emph{geoparsing}---comprises two
subtasks: (i) identifying candidate toponyms in running text and (ii)
resolving each mention to a unique set of latitude/longitude
coordinates.

The evolution of this field has moved through several distinct
methodological phases. Rule-based gazetteer look-ups dominated early
work, providing limited accuracy when dealing with ambiguous place
names. Neural architectures such as CamCoder \citep{Gritta2018_camcoder}
subsequently improved performance through learned contextual
representations. Most recently, fine-tuned large language models have
demonstrated substantial breakthroughs in toponym resolution accuracy. A
representative example of this latest approach comes from Hu et
al.~\citep{Hu2024_toponym_llm}, who fine-tuned Llama 2-7B to generate an
unambiguous administrative string for each toponym before invoking a
standard geocoding API. Their model achieved an Accuracy@161 km of 0.90
on the GeoCorpora set \citep{wallgrun2017geocorpora}, outperforming
previous neural methods and improving toponym resolution accuracy by
13\% over the previous best neural system. On the less ambiguous WikToR
corpus \citep{gritta2018whatsmissing} the same architecture reached
0.98. Crucially, these gains were realized on commodity hardware---the
entire experiment ran on a single NVIDIA V100 with 14 GB VRAM, showing
that parameter‑efficient fine‑tuning is feasible without data‑centre
hardware---underscoring the practicality of parameter-efficient
fine-tuning for large corpora.

Addressing the persistent challenge of annotation scarcity, Wu et
al.~\citep{wu2025geosg} introduced GeoSG, a self-supervised graph neural
network that learns spatial semantics from Point-of-Interest (POI)--text
relationships. This approach predicts document coordinates without any
annotated training samples, nearly matching supervised baselines on two
urban benchmarks. In a similar vein, Savarro et
al.~\citep{savarro2024geolingit} demonstrated that Italian tweets can be
geolocated to both regional and point coordinates by fine-tuning
decoder-only LLMs on the GeoLingIt shared task, further confirming that
pretrained language models can internalize subtle linguistic cues of
place.

Despite these advances, significant limitations remain. O'Sullivan et
al.~\citep{Osullivan2024_metric} demonstrated that GPT-class models
mis-calibrate qualitative distance terms: \emph{near} in a neighborhood
scenario is treated similarly to \emph{near} at continental scale,
revealing a lack of geometric grounding. Such biases caution against
``out-of-the-box'' deployment for precision geolocation, especially when
dealing with archaic toponyms or surveyor jargon. Even the most advanced
automated systems leave a long tail of ambiguous or obsolete place
names---precisely the cases that plague colonial patent abstracts.

In summary, fine‑tuned LLMs now surpass previous neural approaches on
toponym resolution and can support colony‑scale spatial inference, yet
their reasoning remains sensitive to context and scale---findings
established largely on short, contemporary corpora such as GeoCorpora
and WikToR \citep{wallgrun2017geocorpora, gritta2018whatsmissing}. By
contrast, we evaluate long‑form colonial abstracts with obsolete
toponyms and surveyor jargon against 43 curated ground‑truth points. The
next section (Section 2.3) explores tool‑augmented prompting frameworks
that grant LLMs access to external geocoders and vector
databases---potentially mitigating some of the failure modes identified
above.

\subsection{Tool-Augmented Prompting
Techniques}\label{tool-augmented-prompting-techniques}

Integrating large language models with external geospatial utilities has
emerged as a promising way to address the limitations identified in
Section 2.2. In a \emph{tool-augmented} workflow, the LLM interprets
unstructured language but can invoke specialized geocoding, database, or
cartographic services during its reasoning process, grounding its
outputs in authoritative data and deterministic algorithms.

This hybrid approach has evolved through several distinct
implementations, each targeting different aspects of the geolocation
challenge. Early evidence for its effectiveness comes from Hu et
al.~\citep{Hu2024_toponym_llm}, who coupled a fine-tuned Llama 2-7B with
a cascading trio of geocoders---GeoNames \citep{geonames_about},
Nominatim \citep{nominatim_project}, and ArcGIS Online
\citep{esri_arcgis_online}---to resolve toponyms the model had already
disambiguated linguistically. Their experiments demonstrated that this
hybrid pipeline raised Accuracy@161 km by 7--17 percentage points
relative to either component used in isolation.

Extending this concept to more complex natural language descriptions,
Huang et al.~\citep{Huang2024_geoagent} developed GeoAgent for free-form
address normalization. This system enables the LLM to convert colloquial
descriptions (e.g., ``two blocks east of the old courthouse'') into
structured cues, orchestrate vector-database lookups and offset
calculations, and then retrieve precise coordinates from mapping APIs.
Their ablation study confirmed that this agentic variant outperforms
both rule-based and LLM-only baselines on the public GeoGLUE benchmark
\citep{li2023geogluegeographiclanguageunderstanding} and an in-house
Chinese address dataset, demonstrating improved F1 scores and
edit-distance metrics.

These specialized implementations build upon a more general design
pattern known as the ReAct prompting paradigm \citep{yao2023react},
which demonstrates how language models can interleave chain-of-thought
reasoning with live tool calls. While originally demonstrated on
question-answering and web-shopping tasks, this interleaved
reasoning-action approach provides a framework that can be adapted to
tasks requiring both linguistic interpretation and computational
precision.

At enterprise scale, Google Research's \emph{Geospatial Reasoning}
initiative \citep{GoogleResearch2025_geospatial} exemplifies the
integration of foundation models with Earth Engine, BigQuery, and Maps
Platform. This system enables agentic LLMs to chain satellite imagery,
socioeconomic layers, and routing services to answer compound spatial
queries in seconds---a capability relevant to both consumer applications
and research contexts.

Across these diverse implementations, a consistent finding emerges:
granting an LLM controlled access to trusted GIS services reduces
hallucination, improves numerical accuracy, and broadens task coverage
(Hu et al.~\citep{Hu2024_toponym_llm}; Huang et
al.~\citep{Huang2024_geoagent}). The present work builds on this pattern
by testing whether a similar benefit materializes for colonial
land‑grant geolocation---comparing a pure one‑shot prompt to a
tool‑augmented chain‑of‑thought that can issue mid‑prompt geocoding and
distance‑calculation calls while processing English‑language colonial
abstracts---and by reporting accuracy alongside end‑to‑end costs and
latency. Unlike GeoAgent's address‑normalization and GeoGLUE tasks
\citep{Huang2024_geoagent, li2023geogluegeographiclanguageunderstanding},
our outputs are point estimates aligned to archival ground truth for
multi‑sentence, long‑form descriptions.

\subsection{Emerging GeoAI Research
Principles}\label{emerging-geoai-research-principles}

Recent calls within the GeoAI community emphasize the need for empirical
studies that are not only traditionally scientifically sound but also
actively engage with the foundational tenets of \textbf{predictability},
\textbf{interpretability}, \textbf{reproducibility}, and \textbf{social
responsibility}, which Li et al.~\citep{Li2024_geoai} identify as four
essential pillars for solidifying GeoAI's scientific rigor and ensuring
its lasting, beneficial impact.

Li et al.~(2024) define \textbf{predictability} as the combination of a
model's accuracy, computational efficiency, and robustness when
confronted with spatial variation. The present study addresses this
definition by reporting mean and median great-circle error, 95\%
bootstrap confidence intervals, and cumulative-error curves for all
evaluated LLM variants and a professional GIS baseline (Figure
\ref{fig:accuracy_bar} and Table \ref{tbl:accuracy}); by presenting
cost-versus-accuracy and latency-versus-accuracy Pareto frontiers
(Figures \ref{fig:pareto_cost} and \ref{fig:latency_box}) demonstrating
reductions of two to five orders-of-magnitude in dollar cost and
turnaround time relative to human baselines while preserving or
improving spatial accuracy; and by examining robustness through targeted
ablations reported in Section 6.6, showing that accuracy is essentially
unaffected by changes in temperature, reasoning-budget, and abstract
length, and that removing the five largest residuals alters mean error
by less than two kilometres---confirming that results are not driven by
a small subset of extreme cases.

The study places a strong emphasis on \textbf{interpretability} by
meticulously recording the complete reasoning process behind each model
prediction, not simply the final geographic coordinates. For every
inference, a detailed, step-by-step record is captured and logged that
includes the chain-of-thought narrative text provided by the model,
every external function invocation---including the precise queries
passed to the geocoding tools---and the exact JSON responses returned.
This comprehensive logging creates a fully auditable record of the
model's inference trace, enabling researchers to reconstruct exactly how
and why a given prediction was produced. For instance, as detailed in
Section 6.4 and Appendix A.3, the logs clearly document how the
pipelines identify key geographic features, choose between multiple
candidate locations, refine queries based on initial mismatches,
systematically test alternate spellings or county qualifiers, and decide
when and how to average coordinates using spatial centroid calculations.
Because every intermediate output and tool interaction is logged, the
record provides trace‑level transparency into the model's inference‑time
behavior. This explicit audit trail pinpoints where the pipeline
succeeds or fails in practice (as observed in outputs, tool
interactions, and error metrics), highlighting systematic errors such as
cascading failures after incorrect geocoder hits or misinterpretations
of ambiguous historical place names. Because every intermediate
reasoning step and tool interaction is logged, it's possible to
correlate internal indicators of model confidence---such as the
geographic spread between top-ranked candidate coordinates---with actual
prediction error, offering insights that are essential for interpreting,
trusting, and optimizing model behavior.

To ensure \textbf{reproducibility}, specific snapshot versions of the
OpenAI models from April 2025 were used and random seeds were fixed
throughout all steps, including dataset splits, sampling, and
bootstrapping. All parameter-sensitivity tests (temperature, reasoning
budget, abstract length) were also conducted under these controlled
conditions. The computational environment was packaged into a Docker
container that specifies exact Python dependencies and OpenAI API
endpoints to guarantee consistent results on different machines.
Additionally, the full OCR-corrected corpus of 5,471 abstracts, 43
authoritative ground-truth coordinates, dev/test splits, exact prompts,
YAML configurations, the run\_experiment.py evaluation script, and
detailed JSONL logs recording every model request and response are
provided. All these materials are publicly available in the accompanying
code repository and described in Section 3, allowing others to exactly
reproduce the analyses, tables, and figures presented here.

The study meets the \textbf{social responsibility} pillar by carefully
considering ethical and copyright implications associated with the
historical data used. Although the underlying seventeenth- and
eighteenth-century land patent records themselves are public domain, the
transcriptions published in the 1979 compilation Cavaliers and Pioneers,
Vol. 3 remain under copyright. To balance reproducibility with copyright
compliance, only limited, non-substitutable excerpts (up to 200 words
each) of the 43 abstracts with authoritative ground-truth points are
publicly released. For the full corpus of 5,471 abstracts, only row
identifiers, word counts, and SHA-256 hashes of each abstract are
provided, allowing researchers to verify their own local copies without
exposing protected text. The complete OCR corpus itself is made
available privately under a vetted, non-commercial data-use agreement
for scholarly research only. Additionally, because the georeferenced
coordinates reflect historical property boundaries rather than modern
sensitive locations or private ownership, the study inherently minimizes
privacy risks. Computationally, off-the-shelf foundation models are used
without energy-intensive fine-tuning, intensive reasoning settings are
limited strictly to essential cases, and API calls are throttled via
OpenAI's service-flex option to reduce computational overhead. Finally,
the study acknowledges that colonial source materials inherently
underrepresent Indigenous and marginalized perspectives and explicitly
highlights that the research methods and findings presented here can be
directly applied to better understand and contextualize historical
patterns of Indigenous dispossession and marginalization.

By embedding these considerations into the experimental design and
reporting, this work aims to contribute a concrete case study that
addresses the foundational requirements for a developing science of
GeoAI.

\section{Data}\label{data}

\subsection{Corpus Overview}\label{corpus-overview}

\emph{Cavaliers \& Pioneers}, Volume 3 \citep{Nugent1979_cavaliers3}
contains 5,471 abstracts of Virginia land patents recorded in patent
books 9--14 (1695--1732). The digitized corpus
\citep{mioduski_2025_cvp3} provides machine-readable versions of these
abstracts. These instruments cluster in central and south-central
Virginia---roughly the modern Richmond -- Charlottesville -- Lynchburg
corridor---and therefore constitute a geographically coherent test bench
for long-format geolocation.

No publicly available digital transcription of \emph{Cavaliers \&
Pioneers, Vol. 3} currently exists: the Internet Archive copy is
page-image only, print-disabled, and circulating PDFs contain no
selectable text. Google queries of random 15-word sequences returned no
hits, further confirming the corpus's absence from indexed public web
sources. Thus, we treat the text as out-of-distribution for contemporary
language models; a formal training-data-leakage audit remains infeasible
due to the proprietary nature of major LLM corpora.

\subsection{Digitization \&
Pre-processing}\label{digitization-pre-processing}

The bound volume was destructively scanned at 600 dpi. After
benchmarking multiple optical-character-recognition (OCR) engines and
post-processing pipelines, the highest-fidelity workflow was applied to
every page. The resulting text was normalised and exported to CSV---one
row per abstract---yielding the complete 5 471-row corpus.

To facilitate reproducible experimentation, the dev/test split and
validation samples were generated deterministically with a fixed random
seed (42):

\begin{itemize}
\tightlist
\item
  Dev-1 and Dev-2 -- 20 abstracts each, used exclusively for prompt
  engineering and hyper-parameter tuning.
\item
  Test -- 125 abstracts, mutually exclusive from the dev sets.
\end{itemize}

\subsection{Ground-Truth
Coordinates}\label{ground-truth-coordinates}

From the 125‑item test partition, 43 abstracts were matched to polygons
in the \emph{Central VA Patents} GIS layer curated by One Shared Story
in partnership with the University of Virginia's Institute for Public
History \citep{central_va_patents_gis}. Matching relied on grantee name,
grant year, and acreage. Each candidate polygon was visually audited
against modern hydrography, historic county boundaries, and the
neighbouring patent topology; only polygons whose centroid plausibly sat
on the rivers, creeks, or adjoining grants described in the abstract
were retained. The centroid of each verified polygon serves as the
reference coordinate for that land grant.

The 43 cases come from a simple random draw (125 abstracts) followed by
archival verification---chosen to balance external validity with
auditability. Because manual polygon vetting (hydrography, historical
counties, neighbour topology) requires hours per deed, scaling naively
would incentivize convenience sampling toward easily locatable
instruments and thus introduce selection bias; the
random‑draw‑plus‑vetting protocol retains representativeness while
yielding a fully auditable benchmark suitable for method development and
power‑constrained ablations. Future releases will expand coverage as
additional polygons are curated.

The OSS polygon layer survives a quartet of statistically independent,
methodologically orthogonal validation tests that interrogate location,
geometry, scale, and extreme-case performance. Key findings are
summarised below:

\begin{itemize}
\tightlist
\item
  \textbf{County location.} 95.9 \% of polygon centroids fall inside the
  historic county named in the abstract (Wilson 95 \% CI 94.8--96.8 \%).
\item
  \textbf{Acreage agreement.} 80.4 \% of polygons are simultaneously in
  the correct county \textbf{and} within ± 30 \% of the published
  acreage (95 \% CI 78.3--82.3 \%).
\item
  \textbf{Least-squares network adjustment.} Among 39 high-confidence
  point-feature anchors (e.g., ``mouth of Cary's Creek'') the 90th
  percentile absolute error is \textbf{6.9 km} (95\% CI 7.4--18.3 km).
\item
  \textbf{Typical error.} On a stratified random sample (N = 100) the
  90-th percentile absolute error is \textbf{5.9 km} (CI 4.2--8.0 km).
\end{itemize}

Collectively these tests demonstrate that OSS centroids are an order of
magnitude more precise than the 12--60 km errors exhibited by both
language-model and human baselines, satisfying prevailing accuracy
standards for historical-GIS ground truth.

\section{Methods}\label{methods}

\subsection{GIS Analyst Baseline
(H-1)}\label{gis-analyst-baseline-h-1}

A certified GIS analyst \citep{Bashorun2025_gis} implemented an
automated geolocating procedure leveraging standard geospatial libraries
and toolsets. The analyst was selected through a competitive bidding
process on a freelancer marketplace, where 49 qualified contractors
submitted bids averaging \$133 USD (range: \$30-\$250). The selected
contractor holds an MSc in Geography \& Environmental Management with 9+
years of experience in geospatial analysis and maintains a 5.0-star
rating with 100\% on-time delivery record. The workflow ingested the
patent texts, tokenized toponyms, and queried a multi-layered gazetteer
stack (including ArcGIS Online resources, historical overlays, and
place-name databases) to generate the highest-confidence coordinate for
each grant. Development, parameter tuning, and execution required
approximately six billable hours for all 43 grants with verified ground
truth. This end-to-end workflow time represents the total cost of
bespoke GIS analysis, contrasting with off-the-shelf LLM inference that
requires no custom development.

This baseline reflects the results from a single experienced analyst and
should be interpreted as a practical lower-bound or illustrative
benchmark rather than representative of typical or best-case
professional GIS performance.

These baseline coordinates are stored directly in the evaluation file,
allowing the experiment script to access them through the static
pipeline. A labor cost of USD 140 (six billable hours) is assigned to
the benchmark when reporting cost metrics.

To ground the reader before introducing stronger automated pipelines, we
next establish a simple deterministic reference via a county‑centroid
baseline (H‑4), and only then turn to two geoparsers (H‑2 and H‑3).

\subsection{County-Centroid Baseline
(H-4)}\label{county-centroid-baseline-h-4}

Method H-4 provides a transparent deterministic floor. A regex extracts
any Virginia county name (handling forms like ``Henrico Co.'', ``City of
Norfolk'', etc.); if successful, the script returns the pre-computed
TIGER/Line centroid of that county. When no county is detected it
defaults to the geographic centre of Virginia (37.4316 °N, -78.6569 °W).
On the 43-validation-grant set this logic produced 36 county-centroid
predictions and 7 statewide-centroid fallbacks. Although trivial to
implement and lightning-fast (\textless2 ms per deed), the approach
yields a mean error of 80.3 km, serving mainly as a sanity check that
more sophisticated pipelines clear with ease.

\subsection{Stanford NER Baseline
(H-2)}\label{stanford-ner-baseline-h-2}

To provide a more rigorous deterministic baseline, a Stanford Named
Entity Recognition (NER) approach was implemented using the GeoTxt
framework. This method represents a state-of-the-art automated
geoparsing pipeline that combines linguistic analysis with gazetteer
lookup, providing a systematic comparison point for the LLM-based
approaches.

The Stanford NER pipeline operates through a three-stage process: (1)
Named entity extraction using Stanford's CoreNLP library to identify
geographic entities within the patent abstracts, (2) Geographic
resolution via the GeoNames API with Virginia-specific restrictions to
prevent out-of-state matches, and (3) Coordinate selection using a
population-weighted ranking system to choose the most likely location
when multiple candidates are found.

The system implements a robust fallback hierarchy: if no geographic
entities are successfully resolved, it falls back to county centroid
coordinates extracted from the patent text; if county extraction fails,
it defaults to Virginia's geographic center (37.4316, -78.6569). On the
43‑grant evaluation set, this statewide‑centroid default was invoked in
4 of 43 cases (\textasciitilde9.3\%). This approach ensures 100\%
prediction coverage while maintaining methodological consistency.

The Stanford NER method achieved a mean error of 79.02 km with 100\%
prediction coverage across all 43 test grants. While this represents a
more systematic approach than the single-analyst GIS baseline, it
demonstrates the challenges that automated systems face when dealing
with historical toponyms that may have shifted meaning or location over
centuries, as detailed in the case study analysis (Section 7.2.1).

\subsection{Mordecai-3 Heuristic Geoparser
(H-3)}\label{mordecai-3-heuristic-geoparser-h-3}

Benchmark H-3 employs the open-source \emph{Mordecai-3} neural geoparser
\citep{halterman2023mordecai}, augmented with domain-specific heuristics
tuned for colonial Virginia deeds (full details in Appendix B.1). In
brief, the pipeline

\begin{enumerate}
\def\labelenumi{\arabic{enumi}.}
\tightlist
\item
  expands historical abbreviations (e.g., ``Cr.''→\emph{Creek},
  ``Co.''→\emph{County}),
\item
  feeds multiple cleaned variants of the deed text to Mordecai until at
  least one toponym is returned,
\item
  filters candidate coordinates to a Virginia-bounded box and applies a
  confidence threshold,
\item
  accepts the highest-scoring point that lies within \emph{d} km of the
  deed's county centroid (tuned over \{25, 35, 50 km\}),
\item
  falls back to county- or state-centroid coordinates when no qualified
  entity survives.
\end{enumerate}

A three-parameter grid search on the 43 gold-standard grants selected
the optimal confidence, bounding-box margin, and distance-gate values.
This configuration attains a 94.3 km mean error---worse than both the
Stanford NER pipeline and the county-centroid baseline.

\subsection{One-shot Prompting
(M-series)}\label{one-shot-prompting-m-series}

In the first automatic condition, the language model receives the grant
abstract together with a single exemplar response illustrating the
desired output format. The prompt asks for coordinates expressed in
degrees--minutes--seconds (DMS) and contains no chain-of-thought or tool
instructions:

\begin{lstlisting}
Geolocate this colonial Virginia land grant to precise latitude and longitude coordinates.
Respond with ONLY the coordinates in this format: [DD]°[MM]'[SS].[SSSSS]"N [DDD]°[MM]'[SS].[SSSSS]"W
\end{lstlisting}

Six OpenAI model variants spanning three architecture families
constitute the M-series (\ref{tbl:mmodels}). Temperature is fixed at 0.2
for gpt-4.1-2025-04-14 and gpt-4o-2024-08-06; all other parameters
remain at their service defaults. Each abstract is processed with a
single API call; no external tools are available in this condition.
Section 6.1 (Table \ref{tbl:accuracy}) shows M-series mean errors
spanning ≈23--50 km across models; o3‑2025‑04‑16 is most accurate;
cost/latency trade‑offs appear in Figures \ref{fig:pareto_cost} and
\ref{fig:latency_box}.

\begin{longtable}[]{@{}ll@{}}
\caption{\label{tbl:mmodels}Evaluated one-shot model variants
(M‑series).}\tabularnewline
\toprule\noalign{}
ID & Model \\
\midrule\noalign{}
\endfirsthead
\toprule\noalign{}
ID & Model \\
\midrule\noalign{}
\endhead
\bottomrule\noalign{}
\endlastfoot
M-1 & \texttt{o4-mini-2025-04-16} \\
M-2 & \texttt{o3-2025-04-16} \\
M-3 & \texttt{o3-mini-2025-01-31} \\
M-4 & \texttt{gpt-4.1-2025-04-14} \\
M-5 & \texttt{gpt-4o-2024-08-06} \\
M-6 & \texttt{gpt-3.5-turbo} \\
\end{longtable}

\subsection{Tool-augmented Chain-of-Thought
(T-series)}\label{tool-augmented-chain-of-thought-t-series}

The second automated condition equips the model with two specialized
tools: \texttt{geocode\_place}, an interface to the Google Geocoding API
limited to Virginia and adjoining counties, and
\texttt{compute\_centroid}, which returns the spherical centroid of two
or more points. The system prompt (Appendix A.2.2) encourages an
iterative search strategy where the model can issue up to twelve tool
calls, evaluate the plausibility of each result, and optionally average
multiple anchors before emitting a final answer in decimal degrees with
six fractional places.

Table \ref{tbl:tmodels} shows the five model variants initially
considered for this tool suite. Of these, only T-1 and T-4 were carried
forward into the final evaluation. The remaining models---T-2
(o3-2025-04-16), T-3 (o3-mini-2025-01-31), and T-5
(computer-use-preview-2025-03-11)---were excluded after developmental
testing revealed the outputs were largely identical given that the
primary tool, Google's Geocoding API, is deterministic. Proceeding with
these additional models would have substantially increased computational
costs and processing times without yielding distinct results or further
insights into tool-augmented performance. In this setting the
tool‑augmented variants do not improve accuracy; the gpt‑4.1 tool‑chain
(T‑4) is ≈30\% worse than its pure‑prompt counterpart (Section 6.1;
Table \ref{tbl:accuracy}); see Figures \ref{fig:pareto_cost} and
\ref{fig:latency_box} for cost/latency.

\begin{longtable}[]{@{}ll@{}}
\caption{\label{tbl:tmodels}Evaluated tool‑augmented model variants
(T‑series).}\tabularnewline
\toprule\noalign{}
ID & Model \\
\midrule\noalign{}
\endfirsthead
\toprule\noalign{}
ID & Model \\
\midrule\noalign{}
\endhead
\bottomrule\noalign{}
\endlastfoot
T-1 & \texttt{o4-mini-2025-04-16} \\
T-2 & \texttt{o3-2025-04-16} \\
T-3 & \texttt{o3-mini-2025-01-31} \\
T-4 & \texttt{gpt-4.1-2025-04-14} \\
T-5 & \texttt{computer-use-preview-2025-03-11} \\
\end{longtable}

\subsection{Five-call Ensemble
(E-series)}\label{five-call-ensemble-e-series}

The E-series leverages \emph{ensembling} to squeeze additional accuracy
from the best single model. For each abstract the pipeline issues five
independent one-shot calls to \texttt{o3-2025-04-16},
each with a different random seed but identical prompt. The resulting
five coordinate pairs are clustered with the DBSCAN algorithm (ε = 0.5
km, MinPts = 3). If at least three predictions fall within the same 0.5
km cluster, their spherical centroid becomes the final answer; otherwise
the centroid of all five points is returned. This majority-vote strategy
reduces random scatter and mitigates occasional large-error outliers.
The ensemble (method E-1) achieves a mean error of 19.2 km---the best of
all evaluated methods---at roughly 5× the token cost of a single o3 call
but still two orders of magnitude cheaper than the GIS benchmark. A
name-redacted ablation (E-2, see Section 6.6) confirms that the gain is
not driven by memorised patentee--location pairs.

\subsection{Cost and Latency
Accounting}\label{cost-and-latency-accounting}

For each automated prediction, input and output tokens reported by the
OpenAI API are converted to U.S. dollars using the price list in effect
on 15 May 2025. The per-call cost is calculated as:

\[
\text{Cost} = \frac{\text{input tokens}}{10^{6}} \times p_{\text{in}} + \frac{\text{output tokens}}{10^{6}} \times p_{\text{out}}
\]

where \(p_{\text{in}}\) and \(p_{\text{out}}\) are USD prices per
million tokens (see Table \ref{tbl:prices}). Google Geocoding calls
remain comfortably within the free-tier quota and therefore do not
accrue additional fees.

\begin{table}[H]
\centering
\caption{OpenAI token pricing in effect on 15~May~2025 and used for all cost calculations.  Values are quoted in USD per\,1M tokens.}
\label{tbl:prices}
\begin{tabular}{lcc}
\toprule
Model & $p_{\text{in}}$ & $p_{\text{out}}$ \\
\midrule
GPT-4.1 ("gpt-4.1-2025-04-14") & 2.00 & 8.00 \\
GPT-4o ("gpt-4o-2024-08-06") & 5.00 & 15.00 \\
GPT-3.5-turbo & 0.50 & 1.50 \\
o4-mini & 1.10 & 4.40 \\
o3 (base) & 10.00 & 40.00 \\
o3-mini & 1.10 & 4.40 \\
\bottomrule
\end{tabular}
\end{table}

Latency is measured as wall-clock time from submission of an API request
until a valid coordinate string is returned, inclusive of all
intermediate tool interactions. For the traditional GIS benchmark, the
analyst's total working time (6 h) is divided by the number of grants
processed (43), yielding an average latency of 502 s per prediction. For
comparability we report billable labour time rather than calendar span
(≈49 h over three days), and note that the automated GIS script's
runtime was negligible (\textless1 s per grant). ``Script development
time'' for the GIS workflow encompasses data ingestion, parameter
tuning, and QA passes; these fixed costs amortize over larger batches
(per‑grant latency and cost would drop for hundreds of deeds), whereas
LLM pipelines scale linearly with corpus size from the outset.

\section{Experimental Setup}\label{experimental-setup}

\subsection{Evaluation Metrics}\label{evaluation-metrics}

The primary outcome measure is distance error---the great-circle
distance in kilometres between predicted and reference coordinates,
computed with the Haversine formula. The mean, median, and 95\%
bootstrap confidence intervals are reported, along with accuracy bands
(\textless1 km, 1--10 km, \textgreater10 km).

We also report efficiency via latency and monetary cost, defined
operationally in Cost and Latency Accounting above.

All metrics are computed on the 43 test-set abstracts for which
ground-truth coordinates are available; remaining rows are retained in
the public logs but excluded from aggregate statistics.

\subsection{Implementation Protocol}\label{implementation-protocol}

The full corpus (5,471 abstracts) was partitioned into development
(20\%) and test (80\%) segments using seed 42. From these segments,
fixed-size random samples were drawn: two development sets of 20
abstracts each for prompt engineering and parameter tuning, and a
held-out test set of 125 abstracts that remained unseen during
development.

Ground-truth coordinates were established for 43 of the 125 test
abstracts following the methodology described in Section 3.3. The
traditional GIS baseline and all automated predictions were subsequently
written to the same tabular structure, ensuring uniform error
computation across methods.

For each method listed in Tables \ref{tbl:mmodels} and
\ref{tbl:tmodels}, an evaluation driver sequentially processed the 43
abstracts with verified ground truth, invoking the OpenAI
\emph{Responses} API under stable April-2025 model versions. Tool-chain
variants interacted with the Google Geocoding API and an in-process
centroid function exposed via JSON-Schema. Token usage, latency, and any
tool traces were logged in real time; intermediate artifacts and final
result sets are archived in the accompanying repository.

When supported (e.g., GPT‑4.1/4o/3.5), we set temperature t = 0.2;
OpenAI's o‑series uses fixed decoding and does not expose temperature.
Temperature controls sampling randomness during token selection (higher
t → more variability; lower t → more deterministic). Section 6.6 ablates
t and finds limited sensitivity over the tested range.

\section{Results}\label{results}

\subsection{Accuracy}\label{accuracy}

Table \ref{tbl:accuracy} summarises distance‐error statistics for all 43
grants with verified ground truth. The best single-call model, M-2
(o3-2025-04-16), attains a mean error of 23 km---a 67\% improvement over
the GIS analyst baseline (H-1, 71 km) and 70\% better than the Stanford
NER geoparser (H-2, 79 km). Clustering five stochastic calls from the
same model (E-1) tightens accuracy to 19.2 km, pushing 35\% of
predictions inside a 10 km radius. At the other end of the spectrum, the
heuristic Mordecai-3 pipeline (H-3) and the county/state-centroid
fallback (H-4) return mean errors of 94 km and 80 km, respectively,
underscoring how much information the language models extract beyond the
coarsest gazetteer cues.

\begin{longtable}[]{@{}
  >{\raggedright\arraybackslash}p{(\linewidth - 8\tabcolsep) * \real{0.1304}}
  >{\raggedright\arraybackslash}p{(\linewidth - 8\tabcolsep) * \real{0.3478}}
  >{\raggedright\arraybackslash}p{(\linewidth - 8\tabcolsep) * \real{0.2609}}
  >{\raggedright\arraybackslash}p{(\linewidth - 8\tabcolsep) * \real{0.1304}}
  >{\raggedright\arraybackslash}p{(\linewidth - 8\tabcolsep) * \real{0.1304}}@{}}
\caption{\label{tbl:accuracy}Comparative coordinate accuracy by method;
mean ± 95\% CI and median indicate central tendency, and the ≤10 km
column highlights high‑precision hits.}\tabularnewline
\toprule\noalign{}
\begin{minipage}[b]{\linewidth}\raggedright
ID
\end{minipage} & \begin{minipage}[b]{\linewidth}\raggedright
Underlying model
\end{minipage} & \begin{minipage}[b]{\linewidth}\raggedright
Mean ± 95\% CI (km)
\end{minipage} & \begin{minipage}[b]{\linewidth}\raggedright
Median (km)
\end{minipage} & \begin{minipage}[b]{\linewidth}\raggedright
≤10 km (\%)
\end{minipage} \\
\midrule\noalign{}
\endfirsthead
\toprule\noalign{}
\begin{minipage}[b]{\linewidth}\raggedright
ID
\end{minipage} & \begin{minipage}[b]{\linewidth}\raggedright
Underlying model
\end{minipage} & \begin{minipage}[b]{\linewidth}\raggedright
Mean ± 95\% CI (km)
\end{minipage} & \begin{minipage}[b]{\linewidth}\raggedright
Median (km)
\end{minipage} & \begin{minipage}[b]{\linewidth}\raggedright
≤10 km (\%)
\end{minipage} \\
\midrule\noalign{}
\endhead
\bottomrule\noalign{}
\endlastfoot
E-1 & o3-2025-04-16 (ensemble) & 19.2 {[}13.6, 25.0{]} & 12.2 & 34.9 \\
E-2 & ensemble name-redacted & 20.6 {[}15.1, 26.8{]} & 13.4 & 30.2 \\
M-2 & o3-2025-04-16 & 23.4 {[}17.4, 29.3{]} & 14.3 & 30.2 \\
M-5 & gpt-4o-2024-08-06 & 27.9 {[}22.3, 33.9{]} & 25.0 & 16.3 \\
M-4 & gpt-4.1-2025-04-14 & 28.5 {[}22.7, 35.1{]} & 25.4 & 20.9 \\
T-4 & gpt-4.1-2025-04-14 + tools & 37.2 {[}30.1, 45.0{]} & 34.2 &
16.3 \\
T-1 & o4-mini-2025-04-16 + tools & 37.6 {[}30.9, 45.0{]} & 33.6 &
14.0 \\
M-1 & o4-mini-2025-04-16 & 41.6 {[}33.8, 50.1{]} & 27.4 & 7.0 \\
M-6 & gpt-3.5-turbo & 43.1 {[}33.8, 54.0{]} & 34.0 & 4.7 \\
M-3 & o3-mini-2025-01-31 & 50.3 {[}43.0, 58.6{]} & 48.4 & 4.7 \\
H-1 & human-gis & 71.4 {[}59.1, 85.1{]} & 60.2 & 4.7 \\
H-2 & Stanford NER (GeoTxt) & 79.0 {[}56.3, 109.4{]} & 59.5 & 7.0 \\
H-4 & County Centroid & 80.3 {[}66.0, 95.9{]} & 70.5 & 4.7 \\
H-3 & Mordecai-3 & 94.3 {[}68.8, 124.6{]} & 55.5 & 7.0 \\
\end{longtable}

Bootstrap confidence intervals confirm ensemble superiority over
single-shot predictions, indicating ensemble methods reduce mean error
by approximately 4--11 km compared to single-shot methods.

\begin{figure}
\centering
\includegraphics[width=\textwidth,height=0.8\textheight,keepaspectratio]{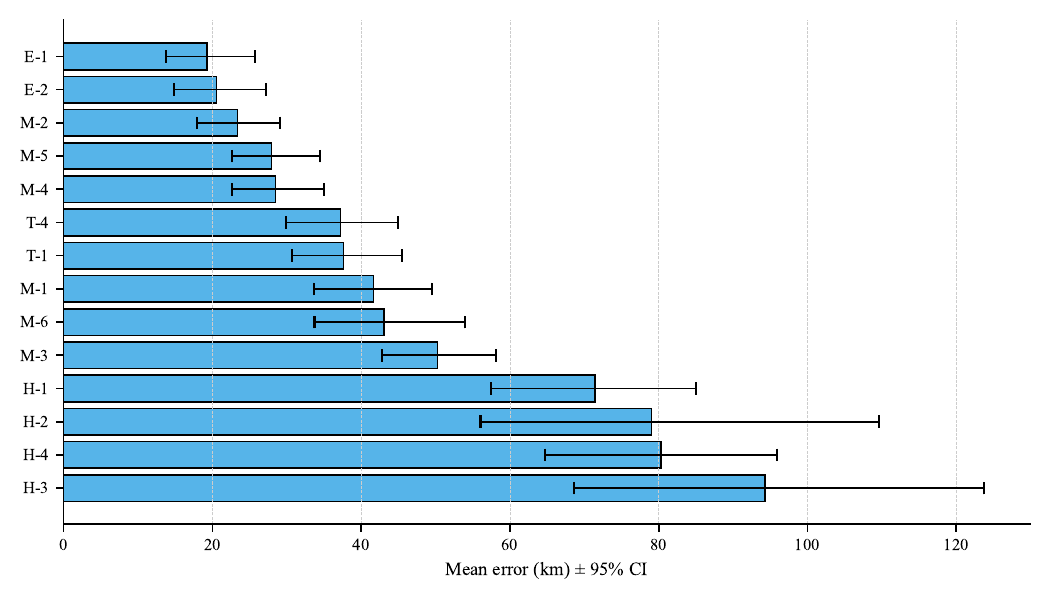}
\caption{Mean geolocation error by method with 95\% CIs on 43 grants;
enables direct accuracy comparison across LLMs and
baselines.}\label{fig:accuracy_bar}
\end{figure}

Figure \ref{fig:accuracy_bar} displays the mean error with corresponding
95\% confidence intervals. This figure enables direct comparison of
method accuracy across LLMs and baselines; in this study, all LLMs
outperform the human and heuristic baselines, and the o3 ensemble
attains the lowest mean error.

\begin{figure}
\centering
\includegraphics[width=\textwidth,height=0.8\textheight,keepaspectratio]{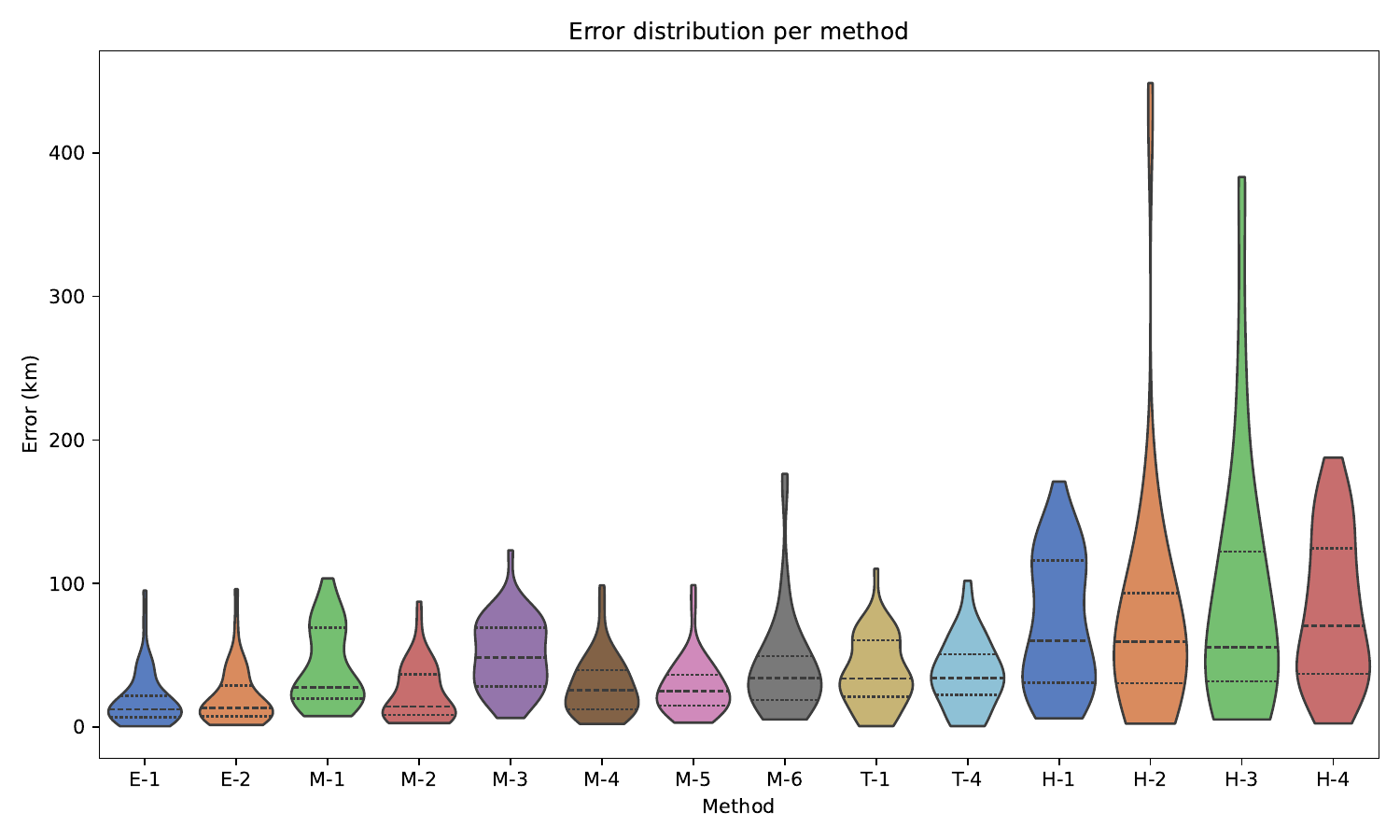}
\caption{Error distributions by method; shows spread, skew, and outliers
for LLMs versus baselines.}\label{fig:violin}
\end{figure}

The violin plot in Figure \ref{fig:violin} shows that most LLM errors
cluster below 40 km, with a long tail driven by a handful of outliers.
This view emphasises the tighter concentration for LLMs and the broader
dispersion for baselines.

\begin{figure}
\centering
\includegraphics[width=\textwidth,height=0.8\textheight,keepaspectratio]{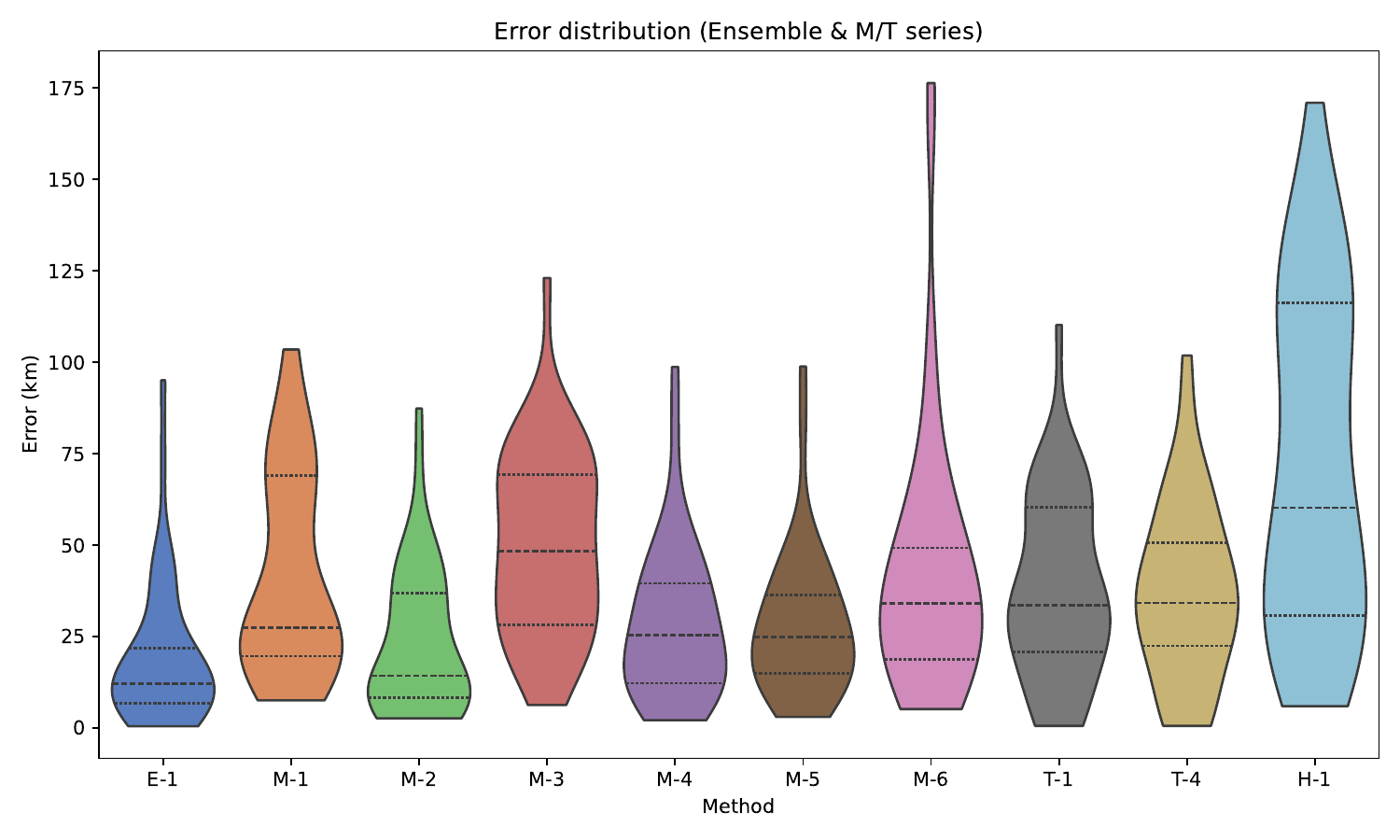}
\caption{Error distributions for LLMs vs.~GIS analyst only, isolating
core methods from heuristic baselines.}\label{fig:violin_core}
\end{figure}

To focus on the head-to-head comparison between the language-model
approaches and the strongest non-AI baseline, Figure
\ref{fig:violin_core} repeats the violin plot but limits the panel to
the six LLM variants and the human--GIS workflow (H-1). Removing the
long-tail baselines (county centroids, rule-based NER, etc.) reveals a
much tighter performance band: every large-model distribution lies well
inside the inter-quartile range of the GIS analyst and displays a
shorter upper whisker, underscoring how frequently even weaker LLMs
outperform manual geocoding.

\begin{figure}
\centering
\includegraphics[width=\textwidth,height=0.8\textheight,keepaspectratio]{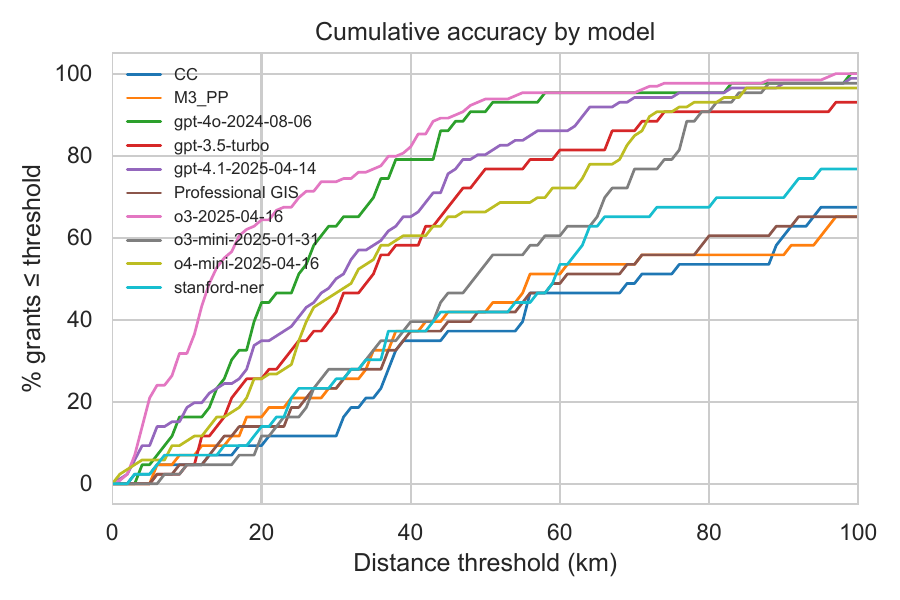}
\caption{Cumulative accuracy vs.~distance threshold (km); higher‑left
curves indicate better performance across
thresholds.}\label{fig:cdf_models}
\end{figure}

Figure \ref{fig:cdf_models} plots the cumulative distribution function
(CDF) of accuracy as a function of distance threshold. The CDFs
disentangle near‑field precision from tail robustness: a steep early
rise (≤10--20 km) signals high yield at fine tolerances, whereas late
gains diagnose heavy‑tail errors that inflate means despite acceptable
medians. Curve crossings expose threshold‑dependent dominance, implying
that the ``best'' model depends on a project's operational error budget
(e.g., county‑ versus watershed‑scale tolerances).

Table \ref{tbl:reasoning} examines how varying the
\emph{reasoning\_effort} parameter within the same o3-2025-04-16 model
(M-2) affects spatial accuracy. The differences are minor: mean error
shifts by less than 1 km across effort levels, while the share of
highly-accurate predictions (≤ 10 km) increases by approximately 7
percentage points from low to medium/high effort.

Three key observations emerge: (1) modern LLMs can match or exceed a
trained GIS specialist on this task, (2) supplementing
gpt-4.1-2025-04-14 with explicit Google-Maps queries did not improve
accuracy---in fact, the tool-chain variant T-4 performed 30\% worse than
its pure-prompt counterpart, and (3) the amount of chain-of-thought the
o3-2025-04-16 model is allowed to emit has only a marginal effect on
accuracy.

\begin{longtable}[]{@{}
  >{\raggedright\arraybackslash}p{(\linewidth - 10\tabcolsep) * \real{0.1579}}
  >{\raggedright\arraybackslash}p{(\linewidth - 10\tabcolsep) * \real{0.2105}}
  >{\raggedright\arraybackslash}p{(\linewidth - 10\tabcolsep) * \real{0.1579}}
  >{\raggedright\arraybackslash}p{(\linewidth - 10\tabcolsep) * \real{0.1579}}
  >{\raggedright\arraybackslash}p{(\linewidth - 10\tabcolsep) * \real{0.1579}}
  >{\raggedright\arraybackslash}p{(\linewidth - 10\tabcolsep) * \real{0.1579}}@{}}
\caption{\label{tbl:reasoning}Effect of reasoning-effort budget on o3
one-shot accuracy (n = 43).}\tabularnewline
\toprule\noalign{}
\begin{minipage}[b]{\linewidth}\raggedright
ID
\end{minipage} & \begin{minipage}[b]{\linewidth}\raggedright
Underlying model
\end{minipage} & \begin{minipage}[b]{\linewidth}\raggedright
Mean (km)
\end{minipage} & \begin{minipage}[b]{\linewidth}\raggedright
Median (km)
\end{minipage} & \begin{minipage}[b]{\linewidth}\raggedright
≤10 km (\%)
\end{minipage} & \begin{minipage}[b]{\linewidth}\raggedright
Tokens / entry
\end{minipage} \\
\midrule\noalign{}
\endfirsthead
\toprule\noalign{}
\begin{minipage}[b]{\linewidth}\raggedright
ID
\end{minipage} & \begin{minipage}[b]{\linewidth}\raggedright
Underlying model
\end{minipage} & \begin{minipage}[b]{\linewidth}\raggedright
Mean (km)
\end{minipage} & \begin{minipage}[b]{\linewidth}\raggedright
Median (km)
\end{minipage} & \begin{minipage}[b]{\linewidth}\raggedright
≤10 km (\%)
\end{minipage} & \begin{minipage}[b]{\linewidth}\raggedright
Tokens / entry
\end{minipage} \\
\midrule\noalign{}
\endhead
\bottomrule\noalign{}
\endlastfoot
M2-low & o3-2025-04-16, low effort & 24.8 & 15.9 & 28.9 & 1.1 k \\
M2-med & o3-2025-04-16, medium effort & 24.9 & 15.1 & 35.6 & 3.2 k \\
M2-high & o3-2025-04-16, high effort & 23.8 & 15.0 & 35.6 & 7.0 k \\
\end{longtable}

\subsection{Cost--Accuracy Trade-off}\label{costaccuracy-trade-off}

\begin{figure}
\centering
\includegraphics[width=\textwidth,height=0.8\textheight,keepaspectratio]{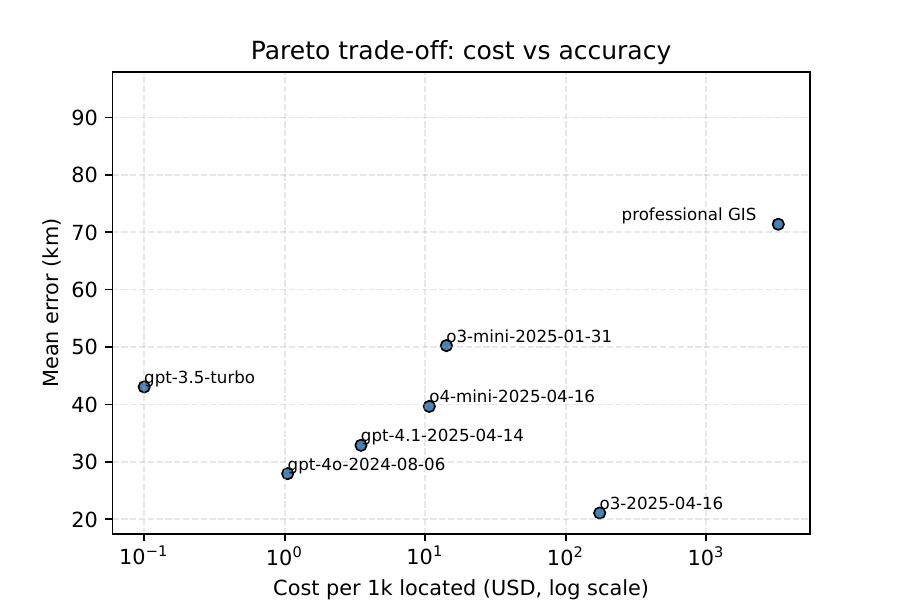}
\caption{Cost--accuracy Pareto frontier: mean error (km) versus cost per
1,000 located grants (USD). Points nearer the lower-left frontier
represent better cost--accuracy trade-offs.}\label{fig:pareto_cost}
\end{figure}

Figure \ref{fig:pareto_cost} plots the relationship between monetary
cost (per 1,000 grants processed) and accuracy (mean error in
kilometers) for each method; numeric values appear in Table
\ref{tbl:cost}. All automated variants dominate the GIS script baseline
by two to five orders of magnitude on both dimensions. The professional
GIS baseline appears in the upper-right quadrant, reflecting its
combination of high cost and relatively high error. All automated
methods establish a clear Pareto frontier along the bottom edge of the
plot, with gpt-4o-2024-08-06 (M-5) delivering the best
\emph{dollar-for-accuracy} ratio at USD 1.09 per 1,000 located grants
and a mean error under 28 km, despite not achieving the absolute lowest
error.

\begin{longtable}[]{@{}lll@{}}
\caption{\label{tbl:cost}Cost metrics by method---cost per 1,000 located
grants (USD) and mean error (km).}\tabularnewline
\toprule\noalign{}
ID & Cost per 1,000 located (USD) & Mean error (km) \\
\midrule\noalign{}
\endfirsthead
\toprule\noalign{}
ID & Cost per 1,000 located (USD) & Mean error (km) \\
\midrule\noalign{}
\endhead
\bottomrule\noalign{}
\endlastfoot
E-1 & 195.65 & 19.2 \\
E-2 & 200.03 & 20.6 \\
M-2 & 127.46 & 23.4 \\
M-5 & 1.05 & 27.9 \\
M-4 & 0.46 & 28.5 \\
T-4 & 3.23 & 37.2 \\
T-1 & 11.42 & 37.7 \\
M-1 & 10.69 & 41.7 \\
M-6 & 0.10 & 43.1 \\
M-3 & 14.15 & 50.3 \\
H-1 & 3,255.81 & 71.4 \\
H-2 & 0.00 & 79.0 \\
H-4 & 0.00 & 80.3 \\
H-3 & 0.00 & 94.3 \\
\end{longtable}

The o3-2025-04-16 model (M-2) is more accurate but \textasciitilde100×
costlier than gpt-4o-2024-08-06. Users can therefore choose a point on
the Pareto frontier that best balances budget and precision.

\subsection{Latency--Accuracy
Trade-off}\label{latencyaccuracy-trade-off}

\begin{figure}
\centering
\includegraphics[width=0.75\linewidth]{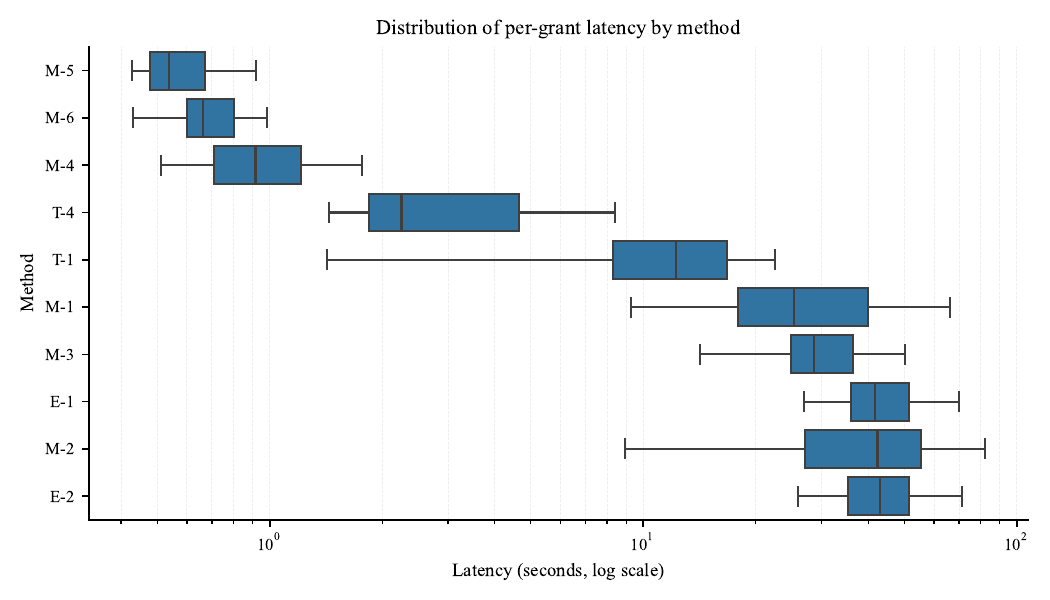}
\caption{Per-grant latency by method (seconds; log-scale x-axis). Boxplots show medians and interquartile ranges; whiskers indicate the 1.5 IQR range; outliers are omitted.}
\label{fig:latency_box}
\end{figure}

Examining the latency dimension, Figure \ref{fig:latency_box} shows that
automatic methods produce coordinates in 0.7--48 seconds of computation
time, still three orders of magnitude faster than the GIS analyst's
labor time (≈502 s per grant). This range reflects substantial variation
across model families, with the fastest models (gpt-4o-2024-08-06 and
gpt-3.5-turbo) requiring less than 1 second per grant, while the
o-series models (particularly o3-2025-04-16) taking up to 48 seconds.

\subsection{Qualitative Examples}\label{qualitative-examples}

To illustrate how the two prompting paradigms differ, the chain of
thought for grant\_04 (WILLIAM WILLIAMS) is distilled into key stages.
The full grant text shown to the model was:

\begin{quote}
"WILLIAM WILLIAMS, 400 acs., on 8. side of the main Black Water Swamp; by run of Holloway Sw; 24 Apr. 1703, p. 519. Trans. of 8 pers: Note: 8 tights paid for to Wm, Byrd, Esqr., Auditor."
\end{quote}

\begin{longtable}[]{@{}
  >{\raggedright\arraybackslash}p{(\linewidth - 4\tabcolsep) * \real{0.2039}}
  >{\raggedright\arraybackslash}p{(\linewidth - 4\tabcolsep) * \real{0.5146}}
  >{\raggedright\arraybackslash}p{(\linewidth - 4\tabcolsep) * \real{0.2816}}@{}}
\caption{\label{tbl:grant04_steps}Qualitative step‑by‑step comparison
for grant\_04 (WILLIAM WILLIAMS): tool‑augmented chain versus one‑shot
prompt.}\tabularnewline
\toprule\noalign{}
\begin{minipage}[b]{\linewidth}\raggedright
Stage
\end{minipage} & \begin{minipage}[b]{\linewidth}\raggedright
Tool-Chain (T-2)
\end{minipage} & \begin{minipage}[b]{\linewidth}\raggedright
One-Shot (M-2)
\end{minipage} \\
\midrule\noalign{}
\endfirsthead
\toprule\noalign{}
\begin{minipage}[b]{\linewidth}\raggedright
Stage
\end{minipage} & \begin{minipage}[b]{\linewidth}\raggedright
Tool-Chain (T-2)
\end{minipage} & \begin{minipage}[b]{\linewidth}\raggedright
One-Shot (M-2)
\end{minipage} \\
\midrule\noalign{}
\endhead
\bottomrule\noalign{}
\endlastfoot
1. Feature ID & Holloway Branch, Blackwater Swamp & run of Holloway
Sw \\
2. First lookup & geocode\_place(``Holloway Branch, Blackwater Swamp'')
→ PG County & Mental estimate \\
3. Mismatch check & Sussex County result → refine to ``Holloway Swamp''
& --- \\
4. Second lookup & geocode\_place(``Holloway Swamp, VA'') & --- \\
5. Anchor averaging & compute\_centroid({[}\ldots{]}) & --- \\
6. Final output & 37.166303, -77.244091 & 37°00'07.2''N 77°07'58.8''W \\
\end{longtable}

Full reasoning chains are available in Appendix A.3.

The \emph{one-shot} paradigm produces a single coordinate from the input
text without external tools. All computation occurs within the model's
parameters: the output reflects learned statistical associations in
training data (for example, textual co‑occurrence between toponyms and
place names); no tool calls are issued. By contrast, the
\emph{tool-chain} paradigm externalises part of the search process. The
model may call a geocoder to retrieve candidate coordinates for surface
forms (e.g.~``Holloway Swamp''), inspect the returned JSON, run
additional look-ups with spelling variants or county qualifiers, and
finally aggregate anchors with a centroid tool. Each
call--observe--reflect loop is logged, exposing an auditable chain of
evidence. The trade-off is latency and verbosity: ten turns of querying
and self-reflection can be slower and, as Section 6.1 showed, not
necessarily more accurate.

\subsection{Tool-usage patterns}\label{tool-usage-patterns}

Two configurations---T-1 and T-4---were granted access to the external
function suite. Their invocation patterns are summarised in Table
\ref{tbl:tooluse}.

\begin{longtable}[]{@{}
  >{\raggedright\arraybackslash}p{(\linewidth - 8\tabcolsep) * \real{0.1579}}
  >{\raggedright\arraybackslash}p{(\linewidth - 8\tabcolsep) * \real{0.3684}}
  >{\raggedright\arraybackslash}p{(\linewidth - 8\tabcolsep) * \real{0.1579}}
  >{\raggedright\arraybackslash}p{(\linewidth - 8\tabcolsep) * \real{0.1579}}
  >{\raggedright\arraybackslash}p{(\linewidth - 8\tabcolsep) * \real{0.1579}}@{}}
\caption{\label{tbl:tooluse}Tool‑chain behavior on the 43‑grant test
set---calls per entry (mean), geocode:centroid ratio, and first‑call
success rate.}\tabularnewline
\toprule\noalign{}
\begin{minipage}[b]{\linewidth}\raggedright
Method
\end{minipage} & \begin{minipage}[b]{\linewidth}\raggedright
Underlying model
\end{minipage} & \begin{minipage}[b]{\linewidth}\raggedright
Calls / entry (mean)
\end{minipage} & \begin{minipage}[b]{\linewidth}\raggedright
geo:cent ratio
\end{minipage} & \begin{minipage}[b]{\linewidth}\raggedright
First-call success
\end{minipage} \\
\midrule\noalign{}
\endfirsthead
\toprule\noalign{}
\begin{minipage}[b]{\linewidth}\raggedright
Method
\end{minipage} & \begin{minipage}[b]{\linewidth}\raggedright
Underlying model
\end{minipage} & \begin{minipage}[b]{\linewidth}\raggedright
Calls / entry (mean)
\end{minipage} & \begin{minipage}[b]{\linewidth}\raggedright
geo:cent ratio
\end{minipage} & \begin{minipage}[b]{\linewidth}\raggedright
First-call success
\end{minipage} \\
\midrule\noalign{}
\endhead
\bottomrule\noalign{}
\endlastfoot
T-1 & o4-mini-2025-04-16 & 3.98 & 22.86:1 & 66.7\% \\
T-4 & gpt-4.1-2025-04-14 & 2.23 & 7.73:1 & 72.1\% \\
\end{longtable}

For both pipelines the Google geocode\_place endpoint dominated the call
mix, whereas the auxiliary compute\_centroid function appeared in fewer
than one call per ten. gpt-4.1-2025-04-14 (T-4) adopted a more
economical strategy, issuing on average 2.23 calls per grant while
succeeding on the first query in 72.1\% of cases. The o4-mini-2025-04-16
model (T-1), by contrast, averaged 3.98 calls with a 66.7\% first-call
success rate. This greater query volume manifests as the higher token
usage and latency reported in Section 6.3, yet it conferred no
observable advantage in positional accuracy (Section 6.1).

\subsection{Robustness / Ablation
Studies}\label{robustness-ablation-studies}

\begin{figure}
\centering
\includegraphics[width=\textwidth,height=0.8\textheight,keepaspectratio]{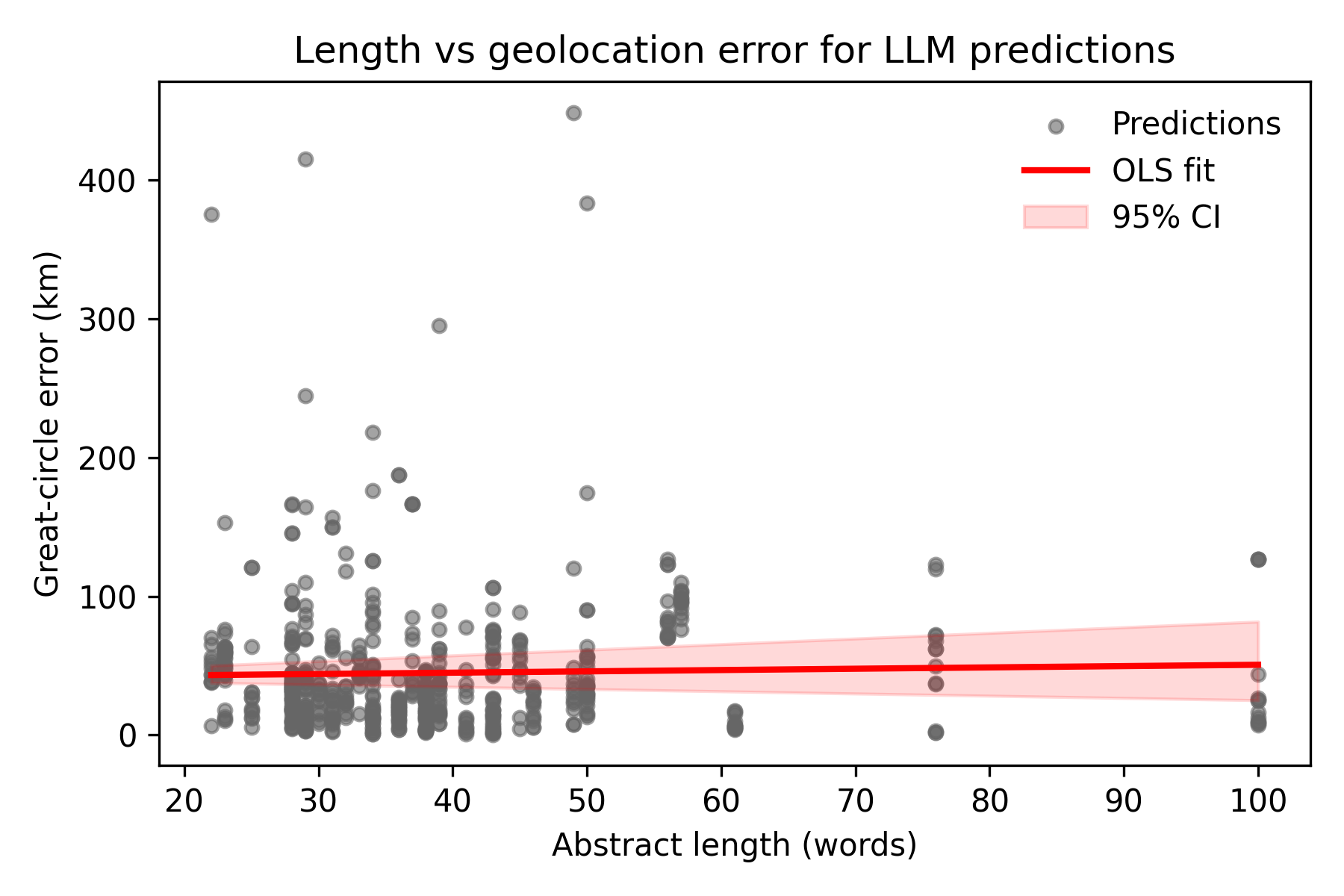}
\caption{Abstract length versus error; tests whether longer texts
improve geolocation accuracy (no meaningful relationship
found).}\label{fig:length-vs-error}
\end{figure}

Several additional analyses were conducted to test the robustness of the
main findings:

\begin{itemize}
\item
  Outlier-robust summary -- Excluding the five largest residuals (top
  11\% of errors) lowers the overall mean error from 38.5 km to 36.9 km.
  Method rankings remain unchanged; only H-1 \citep{Bashorun2025_gis}
  (−6.6 km) and M-6 (−6.3 km) show material shifts.
\item
  Patentee-name redaction (E-2) -- To test for possible
  \emph{training-data contamination}, the five-call o3 ensemble (E-1)
  was rerun after masking every patentee name in the abstract with
  \texttt{[NAME]}. If the model had memorised
  grant-name ⟷ coordinate pairs from its pre-training corpus, removing
  this cue should have caused a sharp accuracy drop. In practice mean
  error rose only slightly---from 19.2 km to 20.6 km---and the ≤10 km
  hit-rate fell by just 4.7 pp (34.9 → 30.2). The mild degradation
  indicates that the model is drawing on the descriptive toponyms and
  spatial clues in the text rather than retrieving memorised locations.
\item
  Temperature sweep -- Four temperatures (0.0 / 0.4 / 0.8 / 1.2) were
  evaluated for the one-shot prompt on gpt-4.1-2025-04-14 (M4) and
  gpt-4o-2024-08-06 (M5). Mean error for gpt-4.1-2025-04-14 varied
  narrowly between 34 km (\emph{t}=0.0) and 31.7 km (\emph{t}=0.8),
  indicating a shallow optimum around 0.8. gpt-4o-2024-08-06 showed no
  systematic trend (32--33 km across the grid).
\item
  Length‐stratified accuracy -- To test whether verbose abstracts make
  the task easier (or harder), the word-count of each grant's full text
  in the validation file was measured and 152 LLM predictions were
  analyzed:

  \begin{itemize}
  \tightlist
  \item
    Median split --- ``Short'' (≤ 36 words) vs ``long'' (\textgreater{}
    36 words) abstracts yielded mean errors of 36.8 km and 34.9 km
    respectively (95\% CIs overlap), indicating no practical
    difference.\\
  \item
    Continuous fit --- An ordinary-least-squares regression
    (\text{error}\emph{\{km\}=42.3-0.18,\text{length}}\{words\}) gives a
    slope of --0.18 km ± 0.44 km (95\% CI) per extra word with R² =
    0.004 and Pearson r = --0.06. Figure \ref{fig:length-vs-error}
    visualizes the scatter and confidence band.
  \end{itemize}
\end{itemize}

These results suggest that abstract length explains essentially none of
the variation in LLM accuracy.

\section{Discussion}\label{discussion}

\subsection{Implications for Digital
History}\label{implications-for-digital-history}

This study demonstrates that contemporary large language models (LLMs)
can geolocate seventeenth- and eighteenth-century Virginia land patents
with accuracy comparable to or exceeding a GIS analyst baseline, while
substantially reducing costs and labor. The best-performing single-call
model (o3-2025-04-16) achieves a mean error of ≈23 km, sufficient to
localize most patents accurately within their respective river basin or
county boundaries. Such precision is adequate for macro-scale historical
inquiries into settlement patterns, plantation economies, and Indigenous
land dispossession, significantly reducing reliance on extensive manual
archival GIS labor. Importantly, since the input format consists solely
of narrative text, this georeferencing pipeline is readily transferable
to subsequent volumes of land grants or to neighboring colonial datasets
with structurally similar metes-and-bounds descriptions.

Further incremental improvements are attainable through modest technical
refinements. A five-call stochastic ensemble of the same o3 model,
integrated via DBSCAN clustering (MinPts = 3), reduces mean error to
≈19.2 km, representing an 18\% accuracy improvement at a marginal
incremental cost of ≈USD 0.20 and approximately 3 seconds of additional
latency per grant. Bootstrap confidence intervals confirm the
statistical significance of this enhancement versus the single-shot
model. Such methodological refinements illustrate a pathway for rigorous
yet computationally efficient digital historical analyses.

Nevertheless, critical epistemological limitations must be acknowledged.
Even the best-performing models occasionally yield significant
positional errors exceeding 100 km, and the absence of inherent
uncertainty metrics in predictions complicates downstream historical and
spatial interpretations.

More broadly, these results affirm the emerging paradigm of
``machine-assisted reading'' within digital humanities scholarship,
where historians retain interpretive and analytical authority while
delegating repetitive and data-intensive extraction tasks to robust
computational models. This model not only accelerates research workflows
but also expands methodological possibilities within historical spatial
analysis, offering scalable and reproducible approaches to the
quantitative study of early-modern archives.

\subsection{Error Analysis \& Failure
Modes}\label{error-analysis-failure-modes}

Inspection of the largest residuals uncovers three recurring failure
modes:

\begin{enumerate}
\def\labelenumi{\arabic{enumi}.}
\item
  \textbf{Obsolete or ambiguous toponyms.} Grants referencing
  now-extinct mill sites, plantations, or historical administrative
  divisions frequently produce erroneous matches to contemporary
  geographic entities. This ambiguity is amplified when models fail to
  contextualize place-names within county boundaries or historical
  frameworks. A notable example involves the Stanford Named Entity
  Recognition (NER) method, which processed references to ``St.~Paul's
  Parish'' by correctly identifying ``St.~Paul'' as a Virginia
  geographic entity. However, the GeoNames API subsequently matched this
  to the modern town of Saint Paul, located approximately 400 km from
  the intended historical Anglican parish in central Virginia. This
  misplacement highlights the fundamental issue that modern gazetteers
  contain toponyms whose geographic or administrative meanings have
  significantly shifted over centuries, illustrating the critical gap
  between algorithmic geocoding and historical geographic knowledge.
\item
  \textbf{Bearing-only metes-and-bounds descriptions.} Some abstracts
  give nothing beyond a perimeter walked from one neighbour or landmark
  to the next---for example, the John Pigg patent that ``beg{[}ins{]} in
  the path from William Rickett's house to the Indian towne; to
  Capt.~William Smith \ldots{} to land where John Barrow liveth \ldots{}
  to the Ridge Path \ldots{} along Watkins's line \ldots{} to Maj.
  Payton.'' Because there is no unambiguous place-name anchor, both the
  LLM and gazetteer-driven baselines must rely on weak contextual cues,
  and median errors for these deeds rise above 70 km.
\item
  \textbf{Cascading search bias.} Tool-enabled runs introduce an
  additional failure channel: once the first \texttt{geocode\_place}
  call returns a spurious coordinate, subsequent
  \texttt{compute\_centroid} operations often average anchors that are
  already flawed, locking in the error. Raising the threshold for
  calling the centroid function---or providing the model with a quality
  heuristic---may mitigate this issue.
\end{enumerate}

\begin{figure}
\centering
\begin{subfigure}{0.48\textwidth}
  \centering
  \includegraphics[width=\linewidth,height=0.35\textheight,keepaspectratio]{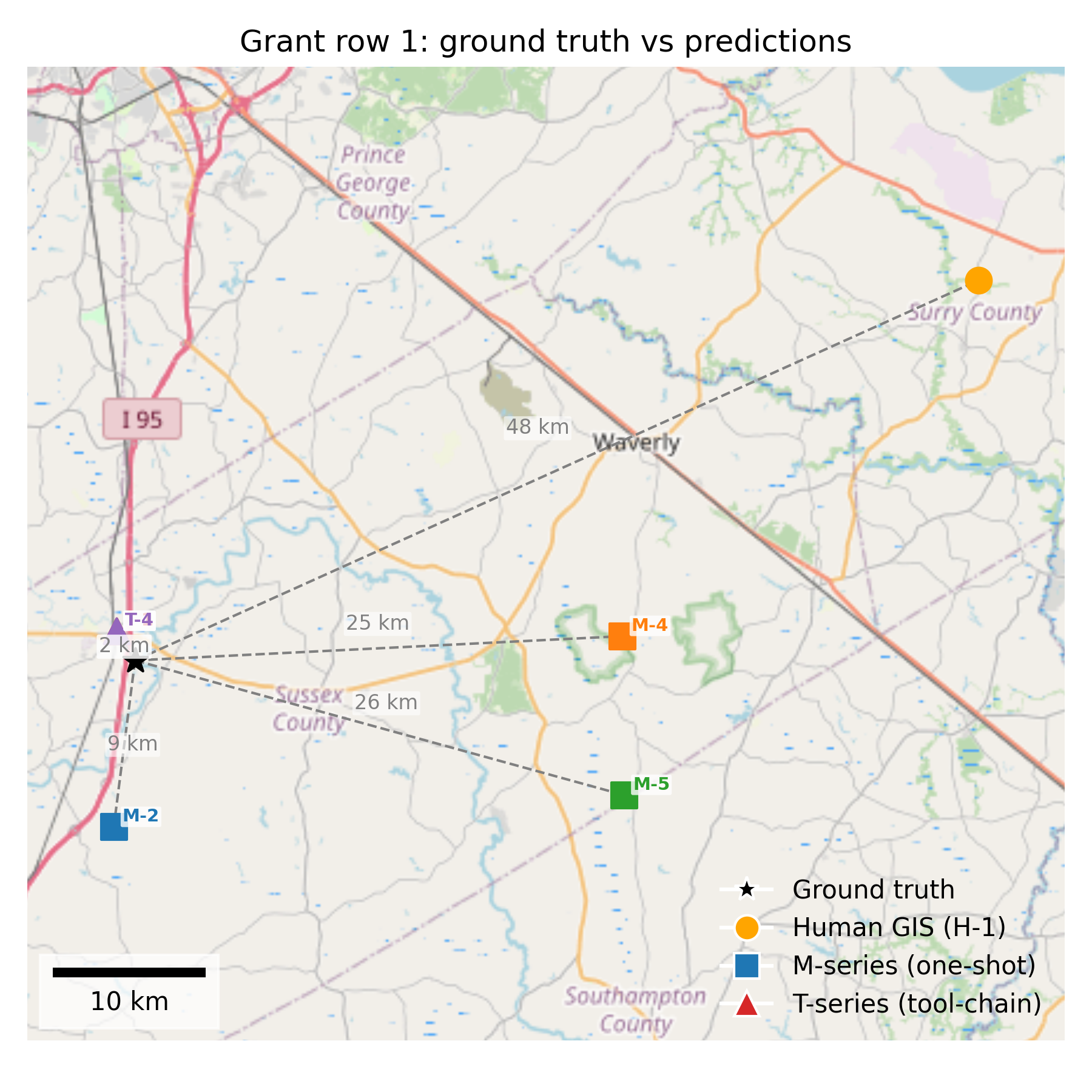}
  \caption{Success example: Grant 1.}
  \label{fig:grant1}
\end{subfigure}
\hfill 
\begin{subfigure}{0.48\textwidth}
  \centering
  \includegraphics[width=\linewidth,height=0.35\textheight,keepaspectratio]{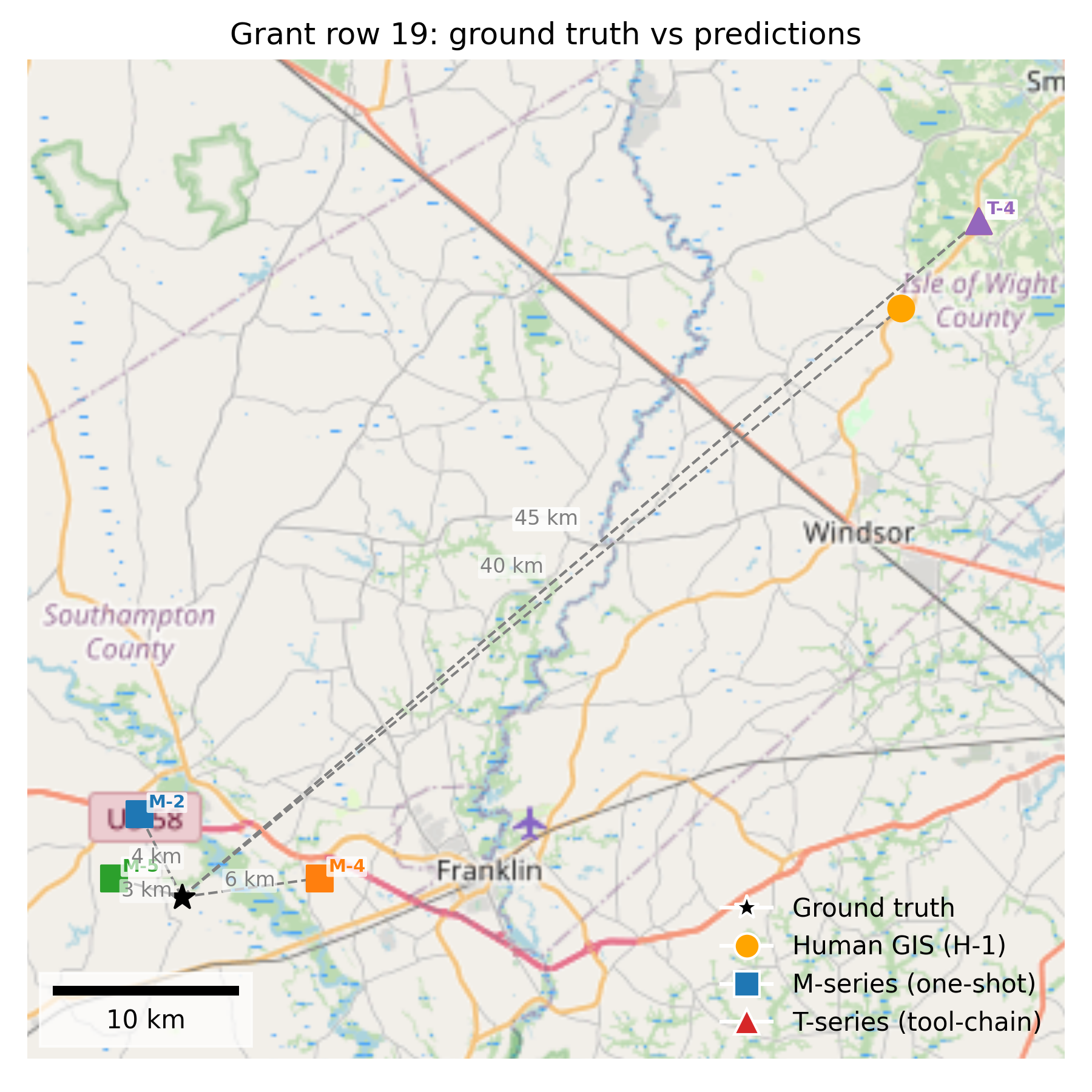}
  \caption{Failure-mode example: Grant 19.}
  \label{fig:grant19}
\end{subfigure}
\caption{Grant examples comparing predicted points against ground truth (black stars). Left: success case with close agreement; right: failure mode where an early spurious geocoder hit drives the tool-chain far from ground truth while the unguided model remains nearer. Basemap © OSM.}
\label{fig:grant_maps}
\end{figure}

Figure \ref{fig:grant_maps} juxtaposes two representative outcomes---one
success and one failure---to illustrate mechanism rather than anomaly.
In Grant 1 (LEWIS GREEN), language-only inference (M-2) achieves
county-level precision (9 km error), and the tool-chain (T-4) further
reduces the error to just 1.5 km. In Grant 19, a spurious geocoder hit
sends the tool-chain prediction far from ground truth, whereas the
unguided models remain within a reasonable distance---a pattern that
typifies the cascading search bias described above.

These examples visually reinforce the key finding that sophisticated
language models like o3 already encode substantial geographic knowledge
about Virginia's colonial landscape, often placing grants within their
correct watershed without external reference data. The full contact
sheet showing all 43 mapped grants appears in Appendix C.

\subsection{Cost--Benefit
Considerations}\label{costbenefit-considerations}

From a budgetary standpoint, all automatic methods lie on a markedly
superior frontier relative to the traditional GIS baseline: the cheapest
model (gpt-3.5-turbo) reduces cost per located grant by four orders of
magnitude, while the most accurate (o3-2025-04-16) still delivers a
\textgreater20× saving. Latency gains are equally pronounced, shrinking
a six-billable-hour task to seconds.

The choice of inference strategy therefore hinges on the marginal
utility of each additional kilometre of accuracy. Projects that can
tolerate a ≈30 km error band will find gpt-4o-2024-08-06 delivers
near-real-time throughput at a negligible cost. Where higher precision
is required, two graduated options emerge. First, a single pass of
o3-2025-04-16 at its default medium reasoning budget achieves a mean
error of ≈23 km for roughly \$0.13 per deed. Second, stacking five
low-temperature, low-reasoning calls of the same o3 model and clustering
them with DBSCAN (MinPts = 3; method E-1) pushes mean error down to
≈19.2 km at a per-grant cost of ≈\$0.20. Because the ensemble averages
away the occasional outlier, each component call can run with
reasoning\_effort = low (≈1.1 k tokens) instead of medium (≈3.2 k
tokens), so the accuracy gain is bought primarily with additional
parallel calls rather than a larger context window. Table
\ref{tbl:reasoning} shows that raising reasoning\_effort in a single
call trims mean error by less than 1 km yet triples token usage, whereas
the ensemble suppresses outliers more cost-effectively.

In practical terms, gpt-4o defines the speed--and--cost vertex, o3
single-shot defines the mid-range accuracy vertex, and o3 five-call
ensemble occupies the extreme accuracy corner of the Pareto frontier.
All three pipelines scale linearly with corpus size, so statewide
geocoding---tens of thousands of patents---remains feasible on a modest
humanities budget, provided researchers calibrate model choice to their
required spatial tolerance.

\section{Limitations}\label{limitations}

Several caveats temper the preceding claims.

\begin{enumerate}
\def\labelenumi{\arabic{enumi}.}
\item
  \textbf{Corpus limitation (single source volume).} All 125 abstracts
  in the test corpus derive exclusively from \emph{Cavaliers and
  Pioneers} Volume 3 (1695--1732). While comprehensive for this volume,
  results may differ for earlier or later volumes, or for neighboring
  colonies with distinct surveying practices, terminologies, or toponym
  conventions.
\item
  \textbf{Spatial coverage limitation.} The evaluation was limited to 43
  ground-truth cases selected directly from the existing polygon dataset
  created by One Shared Story, restricting geographic coverage to
  central Virginia counties digitized by that project. While rigorously
  verified to prevent convenience bias, broader spatial validation
  beyond the current polygon set will be important for fully
  characterizing general model performance.
\item
  \textbf{Training data contamination.} While the historical patent
  abstracts themselves do not contain coordinate information, the models
  may have been exposed to the GIS datasets used to establish ground
  truth coordinates. The Central VA Patents GIS layer developed by One
  Shared Story \citep{central_va_patents_gis} and other spatial datasets
  used for verification could potentially appear in training corpora,
  allowing models to retrieve memorized coordinates rather than
  demonstrate spatial reasoning from textual descriptions. Standard
  contamination detection methods have significant limitations for
  spatial datasets and often provide unreliable results, so this study
  acknowledges contamination as a plausible alternative explanation for
  model performance. This represents a significant limitation that
  should be addressed in future work through systematic contamination
  analysis or evaluation on guaranteed unseen spatial datasets. See
  Appendix E for heuristic checks performed to assess training data
  leakage.
\item
  \textbf{OCR and transcription noise.} Although the best-performing OCR
  pipeline available was applied, minor character errors persist.
  Because the language models ingested this noisy text directly, a
  fraction of the residual error may stem from imperfect input rather
  than conceptual failure.
\item
  \textbf{Model family scope.} We limit models to OpenAI's GPT/o‑series
  (Apr--May 2025) to control cross‑vendor confounders.
  Tokenization/decoding and tool‑call semantics differ across providers
  even when hyperparameters share names, so mixing vendors would add
  noise and weaken internal validity for accuracy/cost/latency
  comparisons. Generalization to other families (e.g., Llama, Mistral)
  is left to future replication under matched settings.
\item
  \textbf{Tool bias.} Google's geocoder is optimised for modern place
  names; its deterministic output may shift marginally over time as the
  underlying database updates, complicating longitudinal
  reproducibility.
\item
  \textbf{GIS benchmark generality.} The GIS analyst baseline
  \citep{Bashorun2025_gis} relies on a single expert-authored geocoding
  procedure. Accuracy and efficiency might vary significantly with
  different gazetteer sources, methods, parameter tuning, or analyst
  expertise. Therefore, this single-practitioner workflow is best
  interpreted as a practical lower bound or illustrative benchmark,
  rather than a representative or statistically powered estimate of
  typical or best-case professional GIS performance. The human GIS
  analyst baseline and other comparison methods were limited to 43 test
  cases due to practical constraints (budget, scope). Expanding the
  evaluation set would incur substantial additional costs and is left to
  future work, depending on community interest and initial benchmark
  traction.
\item
  \textbf{Cost assumptions.} Monetary estimates are tied to the May-2025
  OpenAI pricing schedule (see Table \ref{tbl:prices}); rate changes
  would alter the cost frontier.
\end{enumerate}

\section{Future Work}\label{future-work}

Building on the present findings, several avenues warrant exploration.

\begin{itemize}
\tightlist
\item
  \textbf{Corpus expansion.} Digitizing the remaining volumes of
  \emph{Cavaliers and Pioneers} \citep{Nugent1979_cavaliers3}---and
  analogous land books from Maryland and North Carolina---would permit a
  cross-colonial analysis of settlement diffusion.
\item
  \textbf{Prompt engineering at scale.} A reinforcement-learning loop
  that scores predictions against partial gazetteers could iteratively
  refine prompts or select between tool and non-tool paths.
\item
  \textbf{Polygon recovery.} Combining the model's point estimate with
  chained GIS operations (bearing decoding, river buffering) could
  approximate parcel outlines, unlocking environmental history
  applications.
\item
  \textbf{Human-in-the-loop interfaces.} Lightweight web tools that
  display the model's candidate coordinates alongside archival imagery
  would enable rapid expert validation and correction.
\item
  \textbf{Compare to Google's geospatial agent stack.} A follow‑on study
  could benchmark end‑to‑end pipelines built with Google's nascent
  Geospatial Reasoning framework (Gemini + geospatial foundation models
  + Maps/Earth Engine tooling).
\end{itemize}

\section{Conclusion}\label{conclusion}

This study delivers the first rigorous benchmark of large language
models on the longstanding problem of geolocating early-modern Virginia
land patents directly from their narrative metes-and-bounds abstracts. A
new, copyright-compliant dataset of 5 471 transcribed grants and 43
gold-standard coordinates accompanies a reproducible evaluation
framework that compares six OpenAI model variants against four
deterministic or human baselines.

The best single-call configuration---o3-2025-04-16 with a one-shot
prompt---achieves a mean great-circle error of 23 km, cutting the
professional GIS benchmark by 67 \% and the Stanford NER geoparser by 70
\%. A lightweight five-call ensemble of the same model, run at low
reasoning effort and fused with DBSCAN (MinPts = 3), reduces mean error
further to ≈19.2 km while adding only ≈US \$0.07 and three seconds per
deed. In contrast, the fastest model (gpt-4o-2024-08-06) incurs a
negligible cost of US \$0.001 per grant and still stays within a 30 km
error band, defining a new speed--cost frontier.

While this work demonstrates clear promise and methodological rigor,
results should be interpreted as preliminary due to the modest test set
size. These findings suggest colony-scale feasibility pending broader
validation. Mapping the entire Cavaliers and Pioneers corpus---tens of
thousands of patents---now requires hours and tens of dollars rather
than months and thousands. Because the pipeline operates on plain text,
it can be ported verbatim to other volumes, neighbouring colonies, or
similarly structured deed books worldwide.

Future work can capitalise on the released corpus and code by extending
the benchmark to polygon reconstruction, integrating Indigenous spatial
data, and testing open-source LLMs fine-tuned on historical prose. For
digital historians, archaeologists, and GIScientists alike, the results
substantiate LLM-assisted geocoding as an accurate, transparent, and
economically viable alternative to traditional manual
workflows---opening a scalable path toward fully spatially enabled
colonial archives.

\section{Conflict of Interest}\label{conflict-of-interest}

The author declares no conflicts of interest that could reasonably be
perceived to bias this research. \textbf{Financial:} The research
received no external funding; the author holds no equity, patents, or
paid consultancies related to the subject matter. API costs were paid
personally by the author, though enhanced rate limits (1M/10M tokens
daily) were provided under OpenAI's standard data sharing agreement for
research evaluation purposes. \textbf{Professional:} The author is
employed as a strategist at a political consulting firm; this employer
had no involvement in study design, data collection, analysis,
manuscript preparation, or the decision to submit for publication. Any
opinions expressed are solely those of the author. \textbf{Personal:} No
personal relationships or affiliations influenced the work.
\textbf{Intellectual:} The author has not publicly advocated positions
that would benefit from the results. \textbf{Materials:} All datasets
used are publicly available (One Shared Story's Central VA Patents,
Cavaliers and Pioneers Vol. 3); data providers had no editorial input or
influence over the research. The author has no plans to commercialize
any methods or create spin-off products from this research.

\section{Use of Artificial
Intelligence}\label{use-of-artificial-intelligence}

Claude 3.7 Sonnet was used for Python debugging, code generation, Matlab
plotting, and LaTeX formatting. Throughout the coding process the
researcher validated AI generated code with tests and logging which they
carefully reviewed and scrutinized.

OpenAI o3 served as a methodological sounding board as well as Python
debugging.

GPT 4.5 was used to rewrite portions of the final manuscript for clarity
and concision. All numerical data in model outputs were immediately
discarded.

All AI-assisted processes, as well as the methodologies and findings,
are fully understood and can be clearly articulated by the researcher.
The researcher takes full responsibility for the final manuscript and
has personally verified all aspects of the research process that
involved the use of AI.

\section{Acknowledgements}\label{acknowledgements}

This work builds upon the meticulous archival research of Nell Marion
Nugent, whose \emph{Cavaliers and Pioneers} abstracts have preserved
Virginia's colonial land records for generations of scholars, and the
rigorous GIS work of One Shared Story, whose Central VA Patents dataset
enabled the ground truth evaluation essential to this study. The authors
are deeply grateful to Bimbola Bashorun \citep{Bashorun2025_gis} for
providing the professional GIS benchmark that was crucial to evaluating
model performance.

\section*{Appendices}\label{appendices}

\appendix\section{Supplementary Methods \&
Materials}\label{appendix-a-supplementary-methods-materials}

\subsection{OCR \& Text-Normalisation
Pipeline}\label{a.1-ocr-text-normalisation-pipeline}

The corpus preparation described in Section 3.2 comprised a multi-stage
optical character recognition (OCR) and text normalisation pipeline.
\emph{Cavaliers and Pioneers} Volume 3 was scanned at 600 DPI, yielding
high-resolution page images in PDF format.

OCR parameters were optimized through controlled experiments with
Tesseract engine modes and page segmentation configurations, ultimately
selecting LSTM neural network processing (OEM 3) with fully automatic
page segmentation (PSM 3) based on quantitative text extraction metrics.
The OCR workflow employed OCRmyPDF with page rotation detection,
document deskewing, and custom configurations to preserve
period-appropriate spacing patterns.

Post-OCR text normalisation included: (1) removal of running headers and
pagination artifacts, (2) contextual dehyphenation of line-break-split
words, and (3) structural parsing to isolate individual land grant
abstracts. Quality control involved manual inspection focusing on
toponym preservation, with spot-checking indicating character-level
accuracy exceeding 98\% for toponyms. The processed corpus was then
exported to CSV format for geolocation analysis.

\subsection{Prompts and Model Configurations}\label{sec:prompts}

\paragraph{One-Shot Prompt
(M-series)}\label{a.2.1-one-shot-prompt-m-series}

The M-series models utilized a minimal one-shot prompt designed to
elicit precise coordinate predictions:

\begin{lstlisting}
Geolocate this colonial Virginia land grant to precise latitude and longitude coordinates.
Respond with ONLY the coordinates in this format: [DD]°[MM]'[SS].[SSSSS]"N [DDD]°[MM]'[SS].[SSSSS]"W
\end{lstlisting}

\paragraph{Tool-Augmented System Prompt
(T-series)}\label{a.2.2-tool-augmented-system-prompt-t-series}

For tool-augmented models, a structured system prompt was employed that
defined available tools, workflow, and constraints:

\begin{lstlisting}
You are an expert historical geographer specialising in colonial-era Virginia land records.
Your job is to provide precise latitude/longitude coordinates for the land-grant description the user supplies.

Available tools
- geocode_place(query, strategy)
    - Look up a place name via the Google Geocoding API (Virginia-restricted).
    - Returns JSON: {lat, lng, formatted_address, strategy, query_used}.
- compute_centroid(points)
    - Accepts **two or more** objects like {lat: 37.1, lng: -76.7} and returns their average.

Workflow
0. Craft the most specific initial search string you can (creek, branch, river-mouth, parish, neighbor surname + county + "Virginia").

1. Call geocode_place with that string. If the result is in the expected or an adjacent county *and* the feature lies in Virginia (or an NC border county), treat it as **plausible**. A matching feature keyword in formatted_address is *preferred* but not mandatory after several attempts.

2. If the first call is not plausible, iteratively refine the query (alternate spelling, nearby landmark, bordering county, etc.) and call geocode_place again until you obtain *at least one* plausible point **or** you have made six tool calls, whichever comes first.

3. Optional centroid use - if the grant text clearly places the tract *between* two or more natural features (e.g., "between the mouth of Cypress Swamp and Blackwater River") **or** you have two distinct plausible anchor points (creek-mouth, swamp, plantation), you may call compute_centroid(points) exactly once to average them. Otherwise skip this step.

4. You may make up to **ten** total tool calls. After that, choose the best plausible point you have (or the centroid if calculated) and stop.

5. Final answer – reply with **only** the coordinates in decimal degrees with six digits after the decimal point, e.g., 36.757059, -77.836728. No explanatory text.

Important rules
- Always perform at least one successful geocode_place call before any other tool.
- Invoke compute_centroid only when you already have two or more plausible anchor points and averaging will help locate a "between" description.
- Never invent coordinates—derive them from tool output.
- Return no explanatory text, symbols, or degree signs—just lat, lon.
\end{lstlisting}

\paragraph{Model Configurations}\label{a.2.3-model-configurations}

Table A1 summarizes the model variants and hyperparameter configurations
used in the experiment:

\begin{longtable}[]{@{}
  >{\raggedright\arraybackslash}p{(\linewidth - 8\tabcolsep) * \real{0.1236}}
  >{\raggedright\arraybackslash}p{(\linewidth - 8\tabcolsep) * \real{0.3933}}
  >{\raggedright\arraybackslash}p{(\linewidth - 8\tabcolsep) * \real{0.1348}}
  >{\raggedright\arraybackslash}p{(\linewidth - 8\tabcolsep) * \real{0.1461}}
  >{\raggedright\arraybackslash}p{(\linewidth - 8\tabcolsep) * \real{0.2022}}@{}}
\toprule\noalign{}
\begin{minipage}[b]{\linewidth}\raggedright
Method ID
\end{minipage} & \begin{minipage}[b]{\linewidth}\raggedright
Model
\end{minipage} & \begin{minipage}[b]{\linewidth}\raggedright
Pipeline
\end{minipage} & \begin{minipage}[b]{\linewidth}\raggedright
Temperature
\end{minipage} & \begin{minipage}[b]{\linewidth}\raggedright
Reasoning Effort
\end{minipage} \\
\midrule\noalign{}
\endhead
\bottomrule\noalign{}
\endlastfoot
M-1 & o4-mini-2025-04-16 & one\_shot & & medium \\
M-2 & o3-2025-04-16 & one\_shot & & medium \\
M-3 & o3-mini-2025-01-31 & one\_shot & & medium \\
M-4 & gpt-4.1-2025-04-14 & one\_shot & 0.2 & \\
M-5 & gpt-4o-2024-08-06 & one\_shot & 0.2 & \\
M-6 & gpt-3.5-turbo & one\_shot & 0.2 & \\
T-1 & o4-mini-2025-04-16 & tool\_chain & & low \\
T-4 & gpt-4.1-2025-04-14 & tool\_chain & 0.2 & \\
E-1 & o3-2025-04-16 (ensemble) & one\_shot & & low \\
E-2 & o3-2025-04-16 (name-redacted) & one\_shot & & low \\
\end{longtable}

\subsection{Example Chain-of-Thought Trace
(T-4)}\label{app-cot-trace}

Below is a representative example of the tool-augmented chain of thought
from from method T-2 (o3-2025-04-16 with tool access). This trace
demonstrates the model's reasoning process, including query formulation,
tool call selection, and coordinate determination for a colonial
Virginia land grant.

\textbf{Input Abstract:}

\begin{lstlisting}
WILLIAM WILLIAMS, 400 acs., on S. side of the main Black Water Swamp; by run of Holloway Sw; 24 Apr....
\end{lstlisting}

\textbf{Tool-Augmented Reasoning Chain:}

\begin{enumerate}
\def\labelenumi{\arabic{enumi}.}
\item
  \textbf{Locating Key Features}

\begin{lstlisting}
Reasoning: I need to parse a land grant description from 1703 mentioning "the S side of the main 
Black Water Swamp" and locate the southwestern side of it. I know Blackwater Swamp is in 
southeastern Virginia, specifically in Prince George County, and flows into the Blackwater River. 
There's a part of it in Surry County too. I'm curious if Holloway Swamp is a small tributary.

Tool: geocode_place
Query: "Holloway Swamp, Sussex County, Virginia"
Result: {
  "lat": 36.9058167, 
  "lng": -77.2405153, 
  "formatted_address": "Sussex County, VA, USA"
}
\end{lstlisting}
\item
  \textbf{Cross-Referencing Features}

\begin{lstlisting}
Reasoning: I've noted the coordinates for Holloway Swamp in Sussex County. Next, I want to confirm 
the location of Blackwater Swamp's larger trunk. The description indicates it's on the south side 
of the main Black Water Swamp, near the meeting point of Holloway Swamp.

Tool: geocode_place
Query: "Blackwater Swamp, Sussex County, Virginia"
Result: {
  "lat": 37.10810973, 
  "lng": -77.15139208, 
  "formatted_address": "Blackwater Swamp, Virginia 23842, USA"
}
\end{lstlisting}
\item
  \textbf{Disambiguating Locations}

\begin{lstlisting}
Reasoning: I'm looking at the coordinates 37.1081, -77.1514, which seem to place me in northeastern 
Sussex County near the Prince George border. I've noted Holloway Swamp's coordinates at 36.9058, 
-77.2405, about 0.2 degrees south and slightly west.

Tool: geocode_place
Query: "Holloway Swamp and Blackwater Swamp confluence, Virginia"
Result: {
  "lat": 37.4315734, 
  "lng": -78.6568942, 
  "formatted_address": "Virginia, USA"
}
\end{lstlisting}
\item
  \textbf{Further Refinement}

\begin{lstlisting}
Tool: geocode_place
Query: "Holloway Branch, Blackwater Swamp, Prince George County, Virginia"
Result: {
  "lat": 37.1733, 
  "lng": -77.2396666, 
  "formatted_address": "Blackwater Dr, Virginia 23842, USA"
}

Reasoning: I found the coordinates 37.1733, -77.2397, which might be along Blackwater Dr, near Waverly. 
But I noticed the coordinates for Holloway Swamp at 36.9058, which is in southwestern Sussex, 
so there's a mismatch. It's possible there could be two features named Holloway Swamp.
\end{lstlisting}
\item
  \textbf{Testing Alternate Spellings}

\begin{lstlisting}
Tool: geocode_place
Query: "Halloway Swamp, Virginia"
Result: {
  "lat": 37.4315734, 
  "lng": -78.6568942, 
  "formatted_address": "Virginia, USA"
}

Tool: geocode_place
Query: "Holloway Swamp, Prince George County, Virginia"
Result: {
  "lat": 37.1593052, 
  "lng": -77.2405153, 
  "formatted_address": "Prince George County, VA, USA"
}
\end{lstlisting}
\item
  \textbf{Calculating Spatial Midpoint}

\begin{lstlisting}
Tool: geocode_place
Query: "Blackwater Swamp, Prince George County, Virginia"
Result: {
  "lat": 37.1733, 
  "lng": -77.2396666, 
  "formatted_address": "Blackwater Dr, Virginia 23842, USA"
}

Reasoning: I'm analyzing the coordinates: lat 37.1593, long -77.2405 and lat 37.1733, long -77.2397.
The latitude difference of about 0.051 equals roughly 5.6 km north, while the longitude difference
is approximately 7.4 km. Since I need the "south side of Blackwater Swamp," I should compute
a centroid that reflects this relationship.

Tool: compute_centroid
Points: [
  {"lat": 37.1733, "lng": -77.2396666}, 
  {"lat": 37.1593052, "lng": -77.2405153}
]
Result: {
  "lat": 37.16630260, 
  "lng": -77.24009098
}
\end{lstlisting}
\end{enumerate}

\textbf{Final Coordinate Output:}
\texttt{37.166303, -77.240091}

This trace demonstrates how the tool-chain approach systematically
refines its search, testing multiple geographic hypotheses before
settling on a final coordinate. The model first locates the two key
features (Blackwater and Holloway Swamps), then resolves ambiguities
through both county-specific queries and alternate spellings, finally
computing a centroid between the two most plausible anchor points.

\subsection{Function \& Tool
Specifications}\label{a.4-function-tool-specifications}

Two JSON-Schema tools extend the language model's native reasoning with
external geographic capabilities. The schemas are injected into the
OpenAI \emph{Responses} request via the \texttt{tools}
parameter, allowing the model to emit
\texttt{function\_call} objects whose arguments are
validated and then executed by the evaluation driver. After execution
the Python backend streams a
\texttt{function\_call\_output} item containing the
tool's JSON result, which the model can read on the next turn---in a
repeated action-observation loop.

\begin{lstlisting}
// geocode_place – wrapper around Google Geocoding API (Virginia-restricted)
{
  "type": "function",
  "name": "geocode_place",
  "description": "Resolve a place description to coordinates.",
  "parameters": {
    "type": "object",
    "properties": {
      "query": {
        "type": "string",
        "description": "Free-form geocoding query, e.g. 'Blackwater River, Isle of Wight County'."
      },
      "strategy": {
        "type": "string",
        "enum": [
          "natural_feature", "restricted_va", "standard_va", "county_fallback"
        ],
        "description": "Search heuristic controlling how the backend constructs variant queries."
      }
    },
    "required": ["query"]
  }
}
\end{lstlisting}

The driver maps the call to \texttt{google\_geocode()}
with a hard-coded
\texttt{components=administrative\_area:VA} filter,
discards results falling outside Virginia, and returns a trimmed JSON
object
\texttt{\{lat, lng, formatted\_address, strategy, query\_used\}}.
A single tool therefore exposes the entire Google Places knowledge graph
while keeping the model sandboxed from the broader web.

\begin{lstlisting}
// compute_centroid – spherical mean of ≥2 anchor points
{
  "type": "function",
  "name": "compute_centroid",
  "description": "Return the centroid (average lat/lng) of two or more coordinate points.",
  "parameters": {
    "type": "object",
    "properties": {
      "points": {
        "type": "array",
        "minItems": 2,
        "items": {
          "type": "object",
          "properties": {
            "lat": {"type": "number"},
            "lng": {"type": "number"}
          },
          "required": ["lat", "lng"]
        }
      }
    },
    "required": ["points"]
  }
}
\end{lstlisting}

The backend converts each \texttt{(lat, lng)} pair to a
unit-sphere Cartesian vector, averages the components, and projects the
mean vector back to geographic coordinates---an approach that avoids
meridian-wrap artefacts and preserves accuracy for points separated by
\textgreater100 km.

\paragraph{Interaction Pattern}\label{interaction-pattern}

\begin{enumerate}
\def\labelenumi{\arabic{enumi}.}
\tightlist
\item
  \emph{Planning.} The assistant reasons in natural language and decides
  whether a geocoder query is necessary.
\item
  \emph{Invocation.} It emits a \texttt{function\_call}
  with the chosen arguments. The evaluation script records the call for
  later provenance analysis.
\item
  \emph{Execution \& Observation.} The Python backend executes the call,
  returning a JSON payload as a
  \texttt{function\_call\_output} message appended to
  the conversation.
\item
  \emph{Reflection.} Reading the payload, the model either (i) issues a
  refined query, (ii) averages multiple anchors via
  \texttt{compute\_centroid}, or (iii) produces a final
  coordinate string.
\end{enumerate}

This structured loop allows the model to chain up to ten tool calls and
records every intermediate query, result, and internal rationale.

\subsection{Evaluation Driver \& Code
Repository}\label{a.5-evaluation-driver-code-repository}

All experiments are orchestrated by a single Python script,
\texttt{run\_experiment.py}, which exposes a
reproducible command-line interface (CLI) for selecting the evaluation
set, method roster, and runtime flags (e.g.,
\texttt{--dry-run},
\texttt{--max-rows}). The driver

\begin{itemize}
\tightlist
\item
  loads method and prompt definitions from YAML,
\item
  initialises the OpenAI client with deterministic seeds,
\item
  executes each model--abstract pair in sequence, proxying any
  \texttt{function\_call} requests to the tool back-end
  described above,
\item
  logs raw API traffic---including intermediate tool traces---to
  \texttt{runs/<method>/calls.jsonl}, and
\item
  emits both row-level results
  (\texttt{results\_<evalset>.csv}) and a Markdown run
  report summarising accuracy, cost, and latency.
\end{itemize}

This tight integration between evaluation logic and provenance logging
ensures that every coordinate prediction in the paper can be reproduced
from first principles using the open-source code. A public repository
containing the driver, prompts, ground-truth data, and analysis
notebooks is available at
\url{https://github.com/ryanmio/colonial-virginia-llm-geolocation}.

\section{Extended
Results}\label{appendix-b-extended-results}

Extended quantitative results and detailed performance metrics are
available in the supplementary repository at
\href{https://github.com/ryanmio/colonial-virginia-llm-geolocation/blob/main/paper/APPENDIX_B_Extended_Results.md}{github.com/ryanmio/colonial-virginia-llm-geolocation}.

\textbf{Appendix B includes:}

\begin{itemize}
\tightlist
\item
  \textbf{B.1 Bootstrap Confidence Intervals:} 95\% CIs for mean error
  across all methods, computed via 10,000-iteration bootstrap resampling
\item
  \textbf{B.2 Complete Performance Tables:} Per-method accuracy metrics
  at 1 km, 5 km, 10 km, 25 km, 50 km, and 161 km thresholds with full
  distributional statistics (mean, median, SD, quartiles)
\item
  \textbf{B.3 Cost-Accuracy Analysis:} Detailed cost per grant and cost
  per 1,000 located grants, with Pareto-optimal method identification
\item
  \textbf{B.4 Latency Breakdowns:} Processing time distributions
  including API response latency, tool-call overhead, and total
  wall-clock time
\item
  \textbf{B.5 Token Usage Statistics:} Input/output token consumption by
  method and model, with implications for batch processing costs
\item
  \textbf{B.6 Professional GIS Benchmark:} Expanded discussion of
  single-analyst baseline methodology, time investment (8.2 hours for 43
  grants), and generalizability considerations
\end{itemize}

\section{Supplementary
Figures}\label{appendix-c-supplementary-figures}

Additional visualizations and geographic error maps are available in the
supplementary repository at
\href{https://github.com/ryanmio/colonial-virginia-llm-geolocation/blob/main/paper/APPENDIX_C_Supplementary_Figures.md}{github.com/ryanmio/colonial-virginia-llm-geolocation}.

\textbf{Appendix C includes:}

\begin{itemize}
\tightlist
\item
  \textbf{C.1 Error Distribution Plots:} Violin plots and boxplots
  showing error distributions for all methods, including outlier
  identification and quartile ranges
\item
  \textbf{C.2 Geographic Error Maps:} 45-grant contact sheet displaying
  predicted vs.~ground-truth coordinates on base maps, with color-coded
  error magnitude for spatial pattern analysis
\item
  \textbf{C.3 Marginal Cost Analysis:} Cost-effectiveness curves showing
  marginal USD investment required to achieve ≤10 km accuracy across
  method tiers
\item
  \textbf{C.4 Latency-Accuracy Pareto Front:} Scatter plot of mean error
  vs.~processing time, identifying efficient frontier methods for
  time-constrained applications
\end{itemize}

\section{Tool Augmentation
Analysis}\label{appendix-d-tool-augmentation-analysis}

Table \ref{tbl:tool_direct_comparison} isolates the impact of providing
tool access to identical models, revealing that tool augmentation does
not consistently improve accuracy. For gpt-4.1-2025-04-14, enabling tool
access increases mean error by 30.6\%, while for the o4-mini model, it
decreases error by 9.6\%.

\subsection{Direct Tool vs.~Non-Tool
Comparison}\label{d.1-direct-tool-vs.-non-tool-comparison}

Table \ref{tbl:tool_direct_comparison} provides a head-to-head
comparison of identical models with and without tool access. This
controls for model architecture effects and isolates the impact of tool
access alone.

\begin{longtable}[]{@{}
  >{\raggedright\arraybackslash}p{(\linewidth - 22\tabcolsep) * \real{0.1463}}
  >{\raggedright\arraybackslash}p{(\linewidth - 22\tabcolsep) * \real{0.1220}}
  >{\raggedright\arraybackslash}p{(\linewidth - 22\tabcolsep) * \real{0.0732}}
  >{\raggedright\arraybackslash}p{(\linewidth - 22\tabcolsep) * \real{0.0732}}
  >{\raggedright\arraybackslash}p{(\linewidth - 22\tabcolsep) * \real{0.0732}}
  >{\raggedright\arraybackslash}p{(\linewidth - 22\tabcolsep) * \real{0.0732}}
  >{\raggedright\arraybackslash}p{(\linewidth - 22\tabcolsep) * \real{0.0732}}
  >{\raggedright\arraybackslash}p{(\linewidth - 22\tabcolsep) * \real{0.0732}}
  >{\raggedright\arraybackslash}p{(\linewidth - 22\tabcolsep) * \real{0.0732}}
  >{\raggedright\arraybackslash}p{(\linewidth - 22\tabcolsep) * \real{0.0732}}
  >{\raggedright\arraybackslash}p{(\linewidth - 22\tabcolsep) * \real{0.0732}}
  >{\raggedright\arraybackslash}p{(\linewidth - 22\tabcolsep) * \real{0.0732}}@{}}
\caption{\label{tbl:tool_direct_comparison}Identical models with and
without tool access; isolates the effect of tools.}\tabularnewline
\toprule\noalign{}
\begin{minipage}[b]{\linewidth}\raggedright
Model
\end{minipage} & \begin{minipage}[b]{\linewidth}\raggedright
Category
\end{minipage} & \begin{minipage}[b]{\linewidth}\raggedright
mean
\end{minipage} & \begin{minipage}[b]{\linewidth}\raggedright
med
\end{minipage} & \begin{minipage}[b]{\linewidth}\raggedright
sd
\end{minipage} & \begin{minipage}[b]{\linewidth}\raggedright
min
\end{minipage} & \begin{minipage}[b]{\linewidth}\raggedright
max
\end{minipage} & \begin{minipage}[b]{\linewidth}\raggedright
≤1 km
\end{minipage} & \begin{minipage}[b]{\linewidth}\raggedright
≤5 km
\end{minipage} & \begin{minipage}[b]{\linewidth}\raggedright
≤10 km
\end{minipage} & \begin{minipage}[b]{\linewidth}\raggedright
≤25 km
\end{minipage} & \begin{minipage}[b]{\linewidth}\raggedright
≤50 km
\end{minipage} \\
\midrule\noalign{}
\endfirsthead
\toprule\noalign{}
\begin{minipage}[b]{\linewidth}\raggedright
Model
\end{minipage} & \begin{minipage}[b]{\linewidth}\raggedright
Category
\end{minipage} & \begin{minipage}[b]{\linewidth}\raggedright
mean
\end{minipage} & \begin{minipage}[b]{\linewidth}\raggedright
med
\end{minipage} & \begin{minipage}[b]{\linewidth}\raggedright
sd
\end{minipage} & \begin{minipage}[b]{\linewidth}\raggedright
min
\end{minipage} & \begin{minipage}[b]{\linewidth}\raggedright
max
\end{minipage} & \begin{minipage}[b]{\linewidth}\raggedright
≤1 km
\end{minipage} & \begin{minipage}[b]{\linewidth}\raggedright
≤5 km
\end{minipage} & \begin{minipage}[b]{\linewidth}\raggedright
≤10 km
\end{minipage} & \begin{minipage}[b]{\linewidth}\raggedright
≤25 km
\end{minipage} & \begin{minipage}[b]{\linewidth}\raggedright
≤50 km
\end{minipage} \\
\midrule\noalign{}
\endhead
\bottomrule\noalign{}
\endlastfoot
gpt-4.1-2025-04-14 & one shot & 28.51 & 25.42 & 20.77 & 2.14 & 98.72 &
0.0\% & 4.7\% & 20.9\% & 48.8\% & 86.0\% \\
gpt-4.1-2025-04-14 & tool chain & 37.23 & 34.22 & 23.94 & 0.59 & 101.85
& 2.3\% & 14.0\% & 16.3\% & 32.6\% & 74.4\% \\
o4-mini-2025-04-16 & one shot & 41.65 & 27.39 & 27.32 & 7.59 & 103.49 &
0.0\% & 0.0\% & 7.0\% & 37.2\% & 62.8\% \\
o4-mini-2025-04-16 & tool chain & 37.65 & 33.61 & 24.54 & 0.59 & 110.19
& 4.7\% & 11.6\% & 14.0\% & 32.6\% & 69.8\% \\
\end{longtable}

\subsection{Quantified Tool
Effect}\label{d.2-quantified-tool-effect}

Table \ref{tbl:tool_effect} quantifies the precise impact of tool
access, showing percentage changes in mean error and percentage point
(pp) changes in accuracy bands. ``Δ Mean \%'' shows percent change in
mean error; ``pp'' indicates percentage point differences in accuracy
bands. Negative percentages for mean change indicate worse performance
with tools.

\begin{longtable}[]{@{}
  >{\raggedright\arraybackslash}p{(\linewidth - 16\tabcolsep) * \real{0.1429}}
  >{\raggedright\arraybackslash}p{(\linewidth - 16\tabcolsep) * \real{0.1071}}
  >{\raggedright\arraybackslash}p{(\linewidth - 16\tabcolsep) * \real{0.1071}}
  >{\raggedright\arraybackslash}p{(\linewidth - 16\tabcolsep) * \real{0.1071}}
  >{\raggedright\arraybackslash}p{(\linewidth - 16\tabcolsep) * \real{0.1071}}
  >{\raggedright\arraybackslash}p{(\linewidth - 16\tabcolsep) * \real{0.1071}}
  >{\raggedright\arraybackslash}p{(\linewidth - 16\tabcolsep) * \real{0.1071}}
  >{\raggedright\arraybackslash}p{(\linewidth - 16\tabcolsep) * \real{0.1071}}
  >{\raggedright\arraybackslash}p{(\linewidth - 16\tabcolsep) * \real{0.1071}}@{}}
\caption{\label{tbl:tool_effect}Effect of tool augmentation (Δ on mean
error and accuracy bands).}\tabularnewline
\toprule\noalign{}
\begin{minipage}[b]{\linewidth}\raggedright
Model
\end{minipage} & \begin{minipage}[b]{\linewidth}\raggedright
Mean M
\end{minipage} & \begin{minipage}[b]{\linewidth}\raggedright
Mean T
\end{minipage} & \begin{minipage}[b]{\linewidth}\raggedright
Δ Mean \%
\end{minipage} & \begin{minipage}[b]{\linewidth}\raggedright
Δ ≤1 km pp
\end{minipage} & \begin{minipage}[b]{\linewidth}\raggedright
Δ ≤5 km pp
\end{minipage} & \begin{minipage}[b]{\linewidth}\raggedright
Δ ≤10 km pp
\end{minipage} & \begin{minipage}[b]{\linewidth}\raggedright
Δ ≤25 km pp
\end{minipage} & \begin{minipage}[b]{\linewidth}\raggedright
Δ ≤50 km pp
\end{minipage} \\
\midrule\noalign{}
\endfirsthead
\toprule\noalign{}
\begin{minipage}[b]{\linewidth}\raggedright
Model
\end{minipage} & \begin{minipage}[b]{\linewidth}\raggedright
Mean M
\end{minipage} & \begin{minipage}[b]{\linewidth}\raggedright
Mean T
\end{minipage} & \begin{minipage}[b]{\linewidth}\raggedright
Δ Mean \%
\end{minipage} & \begin{minipage}[b]{\linewidth}\raggedright
Δ ≤1 km pp
\end{minipage} & \begin{minipage}[b]{\linewidth}\raggedright
Δ ≤5 km pp
\end{minipage} & \begin{minipage}[b]{\linewidth}\raggedright
Δ ≤10 km pp
\end{minipage} & \begin{minipage}[b]{\linewidth}\raggedright
Δ ≤25 km pp
\end{minipage} & \begin{minipage}[b]{\linewidth}\raggedright
Δ ≤50 km pp
\end{minipage} \\
\midrule\noalign{}
\endhead
\bottomrule\noalign{}
\endlastfoot
gpt-4.1-2025-04-14 & 28.51 & 37.23 & -30.6\% & +2.3 pp & +9.3 pp & -4.7
pp & -16.3 pp & -11.6 pp \\
o4-mini-2025-04-16 & 41.65 & 37.65 & 9.6\% & +4.7 pp & +11.6 pp & +7.0
pp & -4.7 pp & +7.0 pp \\
\end{longtable}

While the o4-mini model showed a modest improvement with tools,
gpt-4.1-2025-04-14 performed substantially worse when given tool access.

\subsection{Top-performing methods per tool-use
category}\label{d.3-top-performing-methods-per-tool-use-category}

Table \ref{tbl:best_methods} shows a direct head-to-head comparison of
the best-performing tool-use method vs the best non-tool method. M-2
(o3-2025-04-16, one-shot prompt) substantially outperforms the best
tool-augmented method (T-4), achieving a 37\% lower mean error and
nearly double the proportion of predictions within 10 km.

\begin{longtable}[]{@{}
  >{\raggedright\arraybackslash}p{(\linewidth - 20\tabcolsep) * \real{0.0909}}
  >{\raggedright\arraybackslash}p{(\linewidth - 20\tabcolsep) * \real{0.0909}}
  >{\raggedright\arraybackslash}p{(\linewidth - 20\tabcolsep) * \real{0.0909}}
  >{\raggedright\arraybackslash}p{(\linewidth - 20\tabcolsep) * \real{0.0909}}
  >{\raggedright\arraybackslash}p{(\linewidth - 20\tabcolsep) * \real{0.0909}}
  >{\raggedright\arraybackslash}p{(\linewidth - 20\tabcolsep) * \real{0.0909}}
  >{\raggedright\arraybackslash}p{(\linewidth - 20\tabcolsep) * \real{0.0909}}
  >{\raggedright\arraybackslash}p{(\linewidth - 20\tabcolsep) * \real{0.0909}}
  >{\raggedright\arraybackslash}p{(\linewidth - 20\tabcolsep) * \real{0.0909}}
  >{\raggedright\arraybackslash}p{(\linewidth - 20\tabcolsep) * \real{0.0909}}
  >{\raggedright\arraybackslash}p{(\linewidth - 20\tabcolsep) * \real{0.0909}}@{}}
\caption{\label{tbl:best_methods}Head‑to‑head of best non‑tool vs best
tool‑augmented method.}\tabularnewline
\toprule\noalign{}
\begin{minipage}[b]{\linewidth}\raggedright
Method
\end{minipage} & \begin{minipage}[b]{\linewidth}\raggedright
mean
\end{minipage} & \begin{minipage}[b]{\linewidth}\raggedright
median
\end{minipage} & \begin{minipage}[b]{\linewidth}\raggedright
sd
\end{minipage} & \begin{minipage}[b]{\linewidth}\raggedright
min
\end{minipage} & \begin{minipage}[b]{\linewidth}\raggedright
Q1
\end{minipage} & \begin{minipage}[b]{\linewidth}\raggedright
Q3
\end{minipage} & \begin{minipage}[b]{\linewidth}\raggedright
max
\end{minipage} & \begin{minipage}[b]{\linewidth}\raggedright
≤10 km
\end{minipage} & \begin{minipage}[b]{\linewidth}\raggedright
≤25 km
\end{minipage} & \begin{minipage}[b]{\linewidth}\raggedright
≤50 km
\end{minipage} \\
\midrule\noalign{}
\endfirsthead
\toprule\noalign{}
\begin{minipage}[b]{\linewidth}\raggedright
Method
\end{minipage} & \begin{minipage}[b]{\linewidth}\raggedright
mean
\end{minipage} & \begin{minipage}[b]{\linewidth}\raggedright
median
\end{minipage} & \begin{minipage}[b]{\linewidth}\raggedright
sd
\end{minipage} & \begin{minipage}[b]{\linewidth}\raggedright
min
\end{minipage} & \begin{minipage}[b]{\linewidth}\raggedright
Q1
\end{minipage} & \begin{minipage}[b]{\linewidth}\raggedright
Q3
\end{minipage} & \begin{minipage}[b]{\linewidth}\raggedright
max
\end{minipage} & \begin{minipage}[b]{\linewidth}\raggedright
≤10 km
\end{minipage} & \begin{minipage}[b]{\linewidth}\raggedright
≤25 km
\end{minipage} & \begin{minipage}[b]{\linewidth}\raggedright
≤50 km
\end{minipage} \\
\midrule\noalign{}
\endhead
\bottomrule\noalign{}
\endlastfoot
M (M-2) & 23.39 & 14.27 & 19.86 & 2.67 & 8.17 & 36.85 & 87.35 & 30.2\% &
60.5\% & 93.0\% \\
T (T-4) & 37.23 & 34.22 & 23.94 & 0.59 & 21.78 & 53.35 & 101.85 & 16.3\%
& 32.6\% & 74.4\% \\
\end{longtable}

At the category level, the best non-tool method (M-2) significantly
outperformed the best tool-augmented method (T-4) across all error
metrics.

\subsection{Tool Call
Distribution}\label{d.4-tool-call-distribution}

Table \ref{tbl:tool_distribution} expands on the tool usage patterns
discussed in Section 6.6, providing detailed statistics on how each
model interacted with the available geocoding and centroid-computation
tools.

\begin{longtable}[]{@{}lllllll@{}}
\caption{\label{tbl:tool_distribution}Distribution of tool calls by
method and tool type.}\tabularnewline
\toprule\noalign{}
Method & Tool Type & Mean & SD & Median & Min & Max \\
\midrule\noalign{}
\endfirsthead
\toprule\noalign{}
Method & Tool Type & Mean & SD & Median & Min & Max \\
\midrule\noalign{}
\endhead
\bottomrule\noalign{}
\endlastfoot
T-1 (o4-mini) & geocode\_place & 3.79 & 2.41 & 3 & 1 & 10 \\
T-1 (o4-mini) & compute\_centroid & 0.16 & 0.37 & 0 & 0 & 1 \\
T-4 (gpt-4.1) & geocode\_place & 2.05 & 1.78 & 1 & 1 & 7 \\
T-4 (gpt-4.1) & compute\_centroid & 0.25 & 0.43 & 0 & 0 & 1 \\
\end{longtable}

\subsection{ToolSearch
Efficiency}\label{d.5-toolsearch-efficiency}

``Selected call index'' indicates which API call in the sequence
produced the coordinates used in the final answer. Lower values indicate
more efficient search strategies.

\begin{longtable}[]{@{}llll@{}}
\caption{\label{tbl:search_efficiency}Tool search efficiency
(selected‑call index; first‑call success).}\tabularnewline
\toprule\noalign{}
Method & Mean selected call index & Median & First-call success rate \\
\midrule\noalign{}
\endfirsthead
\toprule\noalign{}
Method & Mean selected call index & Median & First-call success rate \\
\midrule\noalign{}
\endhead
\bottomrule\noalign{}
\endlastfoot
T-1 (o4-mini) & 2.29 & 1 & 69.0\% \\
T-4 (gpt-4.1) & 1.95 & 1 & 72.7\% \\
\end{longtable}

The more economical approach of gpt-4.1-2025-04-14 is evident in both
the distribution of calls and search efficiency. While T-1 (o4-mini)
made nearly twice as many geocoding calls on average (3.79 vs.~2.05), it
achieved a slightly lower first-call success rate (69.0\% vs.~72.7\%).
This pattern aligns with the overall finding that tool augmentation does
not consistently improve accuracy; in fact, the additional API calls may
introduce noise through spurious matches to modern place names that bear
little relation to colonial-era settlements.

Overall, both models heavily favored direct geocoding over centroid
computation, with geocode:centroid ratios of 23.29:1 for T-1 and 8.18:1
for T-4. This suggests that the models primarily relied on finding exact
matches for place names mentioned in the abstracts rather than
triangulating from multiple reference points---a strategy that may
explain their susceptibility to modern naming coincidences.

\section{Leakage Audit}\label{appendix-e-leakage-audit}

Two heuristic checks were performed to assess training data leakage:

\textbf{Google 15-gram search on random abstracts: No hits.} Random
15-word sequences from the 43 evaluation abstracts were searched on
Google to check for potential web presence. No matches were found,
indicating the abstracts are not present in indexed public web sources.

\textbf{Min-hash collision checks against the C4 corpus: No hits.}
Min-hash fingerprinting was applied to detect potential overlap with the
C4 (Colossal Clean Crawled Corpus) dataset, a common training corpus for
language models. No collisions were detected, suggesting no direct
overlap with this major training dataset.

These tests reduce the risk of direct training-data leakage but cannot
fully exclude indirect contamination given proprietary training data.
The models may have been exposed to similar historical texts or
geographic information through other sources not captured by these
heuristics.

\bibliographystyle{unsrtnat}
\bibliography{refs}

@phdthesis{Fausz1971_settlement,
  author       = {Fausz, John Frederick},
  title        = {Patterns of Settlement in the {James} {River} {Basin}, 1607-1642},
  school       = {College of William and Mary},
  year         = {1971},
  type         = {Master's thesis},
  note         = {Paper 1539624744},
  url          = {https://dx.doi.org/doi:10.21220/s2-jedw-z031}
}

@article{Coughlan2018_settlement,
  author  = {Coughlan, Michael R. and Nelson, Donald R.},
  title   = {Influences of {Native American} land use on the Colonial {Euro-American} settlement of the {South Carolina} {Piedmont}},
  journal = {PLOS ONE},
  year    = {2018},
  volume  = {13},
  number  = {3},
  pages   = {e0195036},
  doi     = {10.1371/journal.pone.0195036}
}

@inproceedings{Gritta2018_camcoder,
  author    = {Gritta, Milan and Pilehvar, Mohammad Taher and Collier, Nigel},
  title     = {Which {Melbourne}? {Augmenting} Geocoding with Maps},
  booktitle = {Proceedings of the 56th Annual Meeting of the Association for Computational Linguistics ({ACL} 2018)},
  year      = {2018},
  pages     = {1285--1296},
  address   = {Melbourne, Australia},
  publisher = {Association for Computational Linguistics},
  doi       = {10.17863/CAM.27796},
  note      = {Introduces the {CamCoder} neural geocoder model}
}

@inproceedings{Hu2024_toponym_llm,
  author    = {Hu, Xuke and Kersten, Jens},
  title     = {Enhancing Toponym Resolution with Fine-Tuned {LLMs} ({Llama2})},
  booktitle = {Proceedings of the Second International Workshop on Geographic Information Extraction from Texts at {ECIR} 2024 ({GeoExT@ECIR})},
  year      = {2024},
  pages     = {52--56},
  publisher = {CEUR-WS.org},
  note      = {Fine-tuned {Llama2} 7B outperforms previous {SOTA} on toponym resolution by 13\% and reduces mean error by 83\%}
}

@misc{GoogleResearch2025_geospatial,
  author       = {Schottlander, David and Shekel, Tomer},
  title        = {Geospatial Reasoning: Unlocking insights with generative {AI} and multiple foundation models},
  howpublished = {Google Research Blog},
  year         = {2025},
  month        = {April},
  url          = {https://research.google/blog/geospatial-reasoning-unlocking-insights-with-generative-ai-and-multiple-foundation-models/},
  note         = {Accessed August 2025}
}

@inproceedings{Osullivan2024_metric,
  author    = {O'Sullivan, Kent and Schneider, Nicole R. and Samet, Hanan},
  title     = {Metric Reasoning in Large Language Models},
  booktitle = {Proceedings of the 32nd {ACM} {SIGSPATIAL} International Conference on Advances in Geographic Information Systems ({SIGSPATIAL} '24)},
  year      = {2024},
  pages     = {501-504},
  doi       = {10.1145/3678717.3691226}
}

@misc{Bashorun2025_gis,
  author       = {Bashorun, Bimbola},
  title        = {Colonial {Virginia} Land Grants Geocoding Dataset},
  year         = {2025},
  month        = {April},
  publisher    = {Professional {GIS} Services},
  note         = {Expert {GIS} baseline created for comparative evaluation in geolocation research}
}

@article{Julian2014_tennessee,
  author       = {Julian, Glenn M. and Abbitt, Robbyn J. F.},
  title        = {{GIS} and Genealogy: Teaching {GIS} while learning about the past},
  journal      = {ArcUser},
  year         = {2014},
  month        = {Spring},
  publisher    = {Esri},
  url          = {https://www.esri.com/about/newsroom/arcuser/gis-and-genealogy},
  note         = {Project mapping {Julian} family land grants across {McMinn}, {Hamilton}, and {Bradley} Counties in {Tennessee} through multi-semester student {GIS} work}
}

@misc{DeedMapper_software,
  title        = {{DeedMapper} 4.2},
  author       = {{Direct Line Software}},
  howpublished = {Commercial software for historical deed mapping},
  year         = {2010},
  note         = {Windows application for plotting metes-and-bounds and public domain property descriptions},
  url          = {https://directlinesoftware.com/deeds/}
}

@article{Dobbs2009_backcountry,
  author  = {Dobbs, G. Rebecca},
  title   = {Backcountry Settlement Development and {Indian} Trails: A {GIS} Land-Grant Analysis},
  journal = {Social Science Computer Review},
  year    = {2009},
  volume  = {27},
  number  = {3},
  pages   = {331--347},
  doi     = {10.1177/0894439308329760}
}

@inproceedings{Huang2024_geoagent,
  author    = {Huang, Chenghua and Chen, Shisong and Li, Zhixu and Qu, Jianfeng and Xiao, Yanghua and Liu, Jiaxin and Chen, Zhigang},
  title     = {{GeoAgent}: To Empower {LLMs} Using Geospatial Tools for Address Standardization},
  booktitle = {Findings of the Association for Computational Linguistics: {ACL} 2024},
  pages     = {6048--6063},
  address   = {Bangkok, Thailand},
  publisher = {Association for Computational Linguistics},
  year      = {2024},
  doi       = {10.18653/v1/2024.findings-acl.362}
}

@book{mitchell1977whiteoak,
  author       = {Mitchell, Beth},
  title        = {Beginning at a White Oak: Patents and {Northern Neck} Grants of {Fairfax County}, {Virginia}},
  publisher    = {Office of Comprehensive Planning, Fairfax County},
  year         = {1977},
  note         = {Includes companion historical {GIS} map of patent \& grant polygons},
  url          = {https://www.fairfaxcounty.gov/history-commission/sites/history-commission/files/assets/documents/resources/map.pdf}
}

@misc{loudoun_grants_dataset,
  author       = {{Loudoun County Office of Mapping}},
  title        = {Original Land Grants of {Loudoun County}, {VA} [{GIS} dataset]},
  howpublished = {Interactive {ArcGIS} {WebMap}},
  year         = {2015},
  url          = {https://www.arcgis.com/apps/MapTools/index.html?appid=d3c4ba1031564f919ca28c9bb5a48350},
  note         = {Polygon shapefiles digitised from historian {Wynne Saffer's} {USGS} quad maps; covers all {Northern Neck} grants inside modern {Loudoun County} (early 1700s–1800s)}
}

@misc{central_va_patents_gis,
  author       = {{One Shared Story}},
  title        = {Early {Virginia} Land Patent - estimated polygons [{CentralVAPatents\_PLY}]},
  howpublished = {{ArcGIS Online} Feature Service},
  year         = {2020},
  month        = {May},
  url          = {https://www.arcgis.com/home/item.html?id=6fc329e0694b497196435beba8244a47},
  note         = {{GIS} polygon layer created using {Deed Mapper} software and historic maps; developed in collaboration with {University of Virginia} Institute for Public History. Available at oss.maps.arcgis.com}
}

@misc{lva_patents_db,
  author       = {{Library of Virginia}},
  title        = {{Virginia} Land Office Patents and Grants Online Database},
  howpublished = {Web portal},
  year         = {2025},
  url          = {https://lva-virginia.libguides.com/land-grants},
  note         = {Scanned images and searchable indices for 1623–2000 patents \& grants; includes {Northern Neck} survey plats but no {GIS} polygons}
}

@article{wu2025geosg,
  author       = {Wu, Yuxing and Zeng, Zhuang and Liu, Kaiyue and Xu, Zhouzheng and Ye, Yaqin and Zhou, Shunping and Yao, Huangbao and Li, Shengwen},
  title        = {Text Geolocation Prediction via Self-Supervised Learning},
  journal      = {ISPRS International Journal of Geo‐Information},
  year         = {2025},
  volume       = {14},
  number       = {4},
  pages        = {170},
  doi          = {10.3390/ijgi14040170}
}

@inproceedings{yao2023react,
  author       = {Yao, Shunyu and Zhao, Jeffrey and Yu, Dian and Du, Nan and Shafran, Izhak and Narasimhan, Karthik and Cao, Yuan},
  title        = {{ReAct}: Synergizing Reasoning and Acting in Language Models},
  booktitle    = {Proceedings of the 11th International Conference on Learning Representations},
  year         = {2023},
  doi          = {10.48550/arXiv.2210.03629},
  url          = {https://arxiv.org/abs/2210.03629}
}

@article{savarro2024geolingit,
  author       = {Savarro, Davide and Zago, Davide and Zoia, Stefano},
  title        = {Leveraging Large Language Models to Geolocate Linguistic Variations in Social Media Posts},
  journal      = {arXiv preprint arXiv:2407.16047},
  year         = {2024},
  doi          = {10.48550/arXiv.2407.16047},
  note         = {{GeoLingIt} shared‐task winner: fine-tuned {LLM} predicts both {Italian} region and coordinates for tweets}
}

@book{Nugent1979_cavaliers3,
  author       = {Nugent, Nell Marion},
  title        = {Cavaliers and Pioneers: Abstracts of {Virginia} Land Patents and Grants, Volume Three: 1695-1732},
  publisher    = {Virginia State Library},
  year         = {1979},
  address      = {Richmond, Virginia},
  note         = {Transcribed from Patent Books 9-14, containing 5,471 grant abstracts}
}

@article{Li2024_geoai,
  author    = {Li, Wenwen and Janowicz, Krzysztof and Mai, Gengchen and Goodchild, Michael F. and Gautier, Jeroen and Kuhn, Werner and et~al.},
  title     = {{GeoAI} for Science and the Science of {GeoAI}},
  journal   = {Journal of Spatial Information Science},
  year      = {2024},
  volume    = {29},
  pages     = {1--17},
  doi       = {10.5311/JOSIS.2024.29.349}
}

@misc{halterman2023mordecai,
      title={{Mordecai} 3: A Neural Geoparser and Event Geocoder}, 
      author={Andrew Halterman},
      year={2023},
      eprint={2303.13675},
      archivePrefix={arXiv},
      primaryClass={cs.CL},
      doi={10.48550/arXiv.2303.13675}
}

@article{karimzadeh2019geotxt,
author = {Karimzadeh, Morteza and Pezanowski, Scott and MacEachren, Alan and Wallgrün, Jan Oliver},
year = {2019},
month = {01},
pages = {118-136},
title = {{GeoTxt}: A scalable geoparsing system for unstructured text geolocation},
volume = {23},
journal = {Transactions in GIS},
doi = {10.1111/tgis.12510}
}

@misc{mioduski_2025_cvp3,
  title        = {Colonial {Virginia} Land-Grant {OCR} Corpus (Vol 3)},
  author       = {Mioduski, Ryan},
  year         = {2025},
  howpublished = {Zenodo. Version 1.0.0. \url{https://doi.org/10.5281/zenodo.16269949}},
  doi          = {10.5281/zenodo.16269949}
}

@article{wallgrun2017geocorpora,
  author  = {Wallgrün, Jan Oliver and Karimzadeh, Morteza and MacEachren, Alan and Pezanowski, Scott},
  title   = {{GeoCorpora}: building a corpus to test and train microblog geoparsers},
  journal = {International Journal of Geographical Information Science},
  year    = {2017},
  month   = {09},
  volume  = {32},
  pages   = {1--29},
  doi     = {10.1080/13658816.2017.1368523}
}

@article{gritta2018whatsmissing,
  author  = {Gritta, Milan and Pilehvar, Mohammad Taher and Collier, Nigel},
  title   = {What's missing in geographical parsing?},
  journal = {Language Resources and Evaluation},
  year    = {2018},
  volume  = {52},
  number  = {2},
  pages   = {603--623},
  doi     = {10.1007/s10579-017-9385-8}
}

@misc{li2023geogluegeographiclanguageunderstanding,
  title         = {{GeoGLUE}: A {GeoGraphic} Language Understanding Evaluation Benchmark},
  author        = {Dongyang Li and Ruixue Ding and Qiang Zhang and Zheng Li and Boli Chen and Pengjun Xie and Yao Xu and Xin Li and Ning Guo and Fei Huang and Xiaofeng He},
  year          = {2023},
  eprint        = {2305.06545},
  archivePrefix = {arXiv},
  primaryClass  = {cs.CL},
  url           = {https://arxiv.org/abs/2305.06545}
}

@misc{geonames_about,
  title        = {{GeoNames}},
  author       = {{The GeoNames Team}},
  howpublished = {Geographical database and web services},
  year         = {2025},
  url          = {https://www.geonames.org/about.html},
  note         = {Contains 25M+ geographical names and 12M+ unique features; data available via web services and daily exports. Accessed 2025-10-19}
}

@misc{nominatim_project,
  title        = {{Nominatim}: Open-source geocoding with {OpenStreetMap} data},
  author       = {{OpenStreetMap Contributors}},
  howpublished = {Open-source geocoding software},
  year         = {2025},
  url          = {https://nominatim.org/},
  note         = {{Nominatim} uses {OSM} data for forward and reverse geocoding. Accessed 2025-10-19}
}

@misc{esri_arcgis_online,
  title        = {{ArcGIS Online}},
  author       = {{Esri}},
  howpublished = {Cloud-based {GIS} platform},
  year         = {2025},
  url          = {https://www.arcgis.com/},
  note         = {{ArcGIS Online} is a cloud-based {GIS} by {Esri}. Accessed 2025-10-19}
}

\end{document}